\documentclass[10pt,letterpaper]{article}
\usepackage[top=0.85in,left=2.75in,footskip=0.75in]{geometry}

\usepackage[english]{babel}

\usepackage{changepage}

\usepackage[utf8x]{inputenc}

\usepackage{amsmath}
\usepackage{amsfonts} 
\usepackage{amssymb} 
\usepackage{fixmath}
\usepackage{mathtools}
\usepackage{nicefrac}
\mathtoolsset{showonlyrefs}

\usepackage{booktabs}
\usepackage{csquotes}
\usepackage[super]{nth}
\usepackage{url}

\usepackage{longtable}

\usepackage{textcomp,marvosym}

\usepackage{cite}

\usepackage[table]{xcolor}
\usepackage{nameref,hyperref}
\newcommand\myshade{50}
\colorlet{mylinkcolor}{blue}
\colorlet{mycitecolor}{green}
\colorlet{myurlcolor}{red}
\hypersetup{
  linkcolor  = mylinkcolor!\myshade!black,
  citecolor  = mycitecolor!\myshade!black,
  urlcolor   = myurlcolor!\myshade!black,
  colorlinks = true,
}

\usepackage{microtype}
\usepackage{xspace}
\DisableLigatures[f]{encoding = *, family = * }

\usepackage{array}

\newcolumntype{+}{!{\vrule width 2pt}}

\newlength\savedwidth

\raggedright
\usepackage[aboveskip=1pt,labelfont=bf,labelsep=period,justification=raggedright,singlelinecheck=off]{caption}

\makeatletter
\renewcommand{\@biblabel}[1]{\quad#1.}
\makeatother

\usepackage{lastpage,fancyhdr,graphicx}
\graphicspath{{./figures/}}
\usepackage{epstopdf}
\fancyheadoffset[L]{2.25in}
\fancyfootoffset[L]{2.25in}
\def\RR{{\mathbb R}}

\def\x{{\mathbold x}}

\def\w{{\mathbold w}}

\def\l{{\mathbold l}}

\def\A{{\mathbold A}}

\def\L{{\mathbold L}}

\def\D{{\mathbold D}}
\def\T{{\mathbold T}}

\def\I{{\mathbold I}}

\def\L{{\mathbold L}}
\def\M{{\mathbold M}}
\def\K{{\mathbold K}}

\def\U{{\mathbold U}}
\def\V{{\mathbold V}}
\def\X{{\mathbold X}}

\def\p{{\mathbold \p}}

\def\b{{\mathbold b}}

\def\d{{\mathbold d}}

\def\x{{\mathbold x}}
\def\W{{\mathbold W}}

\def\p{{\mathbold p}}

\newcommand{\cC}{{\mathcal C}}

\DeclareMathOperator*{\argmin}{argmin}
\DeclareMathOperator*{\argmax}{argmax}

\newcommand\diag{\text{Diag}}

\definecolor{dred}{rgb}{0.8,0,0}
\definecolor{dgreen}{rgb}{0,0.8,0}
\definecolor{dblue}{rgb}{0,0,0.8}
\definecolor{dpurple}{rgb}{0.8,0,0.8}
\colorlet{darkblue}{blue!80!black}
\colorlet{darkred}{red!80!black}
\colorlet{darkgreen}{green!60!black}
\colorlet{darkmagenta}{orange!80!black}
\colorlet{darkyellow}{purple!80!black}

\newcommand{\ie}{i.e.,\xspace}

\newcommand{\eg}{e.g.,\xspace}

\newcommand{\acs}{\textsc{}}
\newcommand{\acsfont}{\textsc{}}

\makeatletter
\addto\extrasenglish{%
}
\addto\captionsenglish{}
\makeatother

\usepackage{etoolbox}
\makeatletter
\patchcmd{\hyper@makecurrent}{%
    \ifx\Hy@param\Hy@chapterstring
        \let\Hy@param\Hy@chapapp
    \fi
}{%
    \iftoggle{inappendix}{
        \@checkappendixparam{chapter}%
        \@checkappendixparam{section}%
        \@checkappendixparam{subsection}%
        \@checkappendixparam{subsubsection}%
        \@checkappendixparam{paragraph}%
        \@checkappendixparam{subparagraph}%
    }{}%
}{}{\errmessage{failed to patch}}

\newcommand*{\@checkappendixparam}[1]{%
    \def\@checkappendixparamtmp{#1}%
    \ifx\Hy@param\@checkappendixparamtmp
        \let\Hy@param\Hy@appendixstring
    \fi
}
\makeatletter

\newtoggle{inappendix}
\togglefalse{inappendix}

\apptocmd{\appendix}{\toggletrue{inappendix}}{}{\errmessage{failed to patch}}

\begin{document}

\vspace*{0.2in}
\begin{flushleft}
{\Large
\textbf\newline{Extracting representations of cognition across neuroimaging
studies improves brain decoding} 
}
\newline
\\
Arthur Mensch\textsuperscript{1*},
Julien Mairal\textsuperscript{2},
Bertrand Thirion\textsuperscript{1},
Gaël Varoquaux\textsuperscript{1},
\\
\bigskip
\textbf{1} Inria, CEA,  Univ. Paris Saclay, Paris, France
\\
\textbf{2} Univ. Grenoble Alpes, Inria, CNRS,
Grenoble INP, LJK, Grenoble, France
\\
\bigskip

* arthur.mensch@m4x.org

\end{flushleft}

\section*{Abstract}

Cognitive brain imaging is accumulating datasets about the neural substrate of many different mental processes. Yet, most studies are based on few subjects and have low statistical power. Analyzing data across studies could bring more statistical power; yet the current brain-imaging analytic framework cannot be used at scale as it requires casting all cognitive tasks in a unified theoretical framework. We introduce a new methodology to analyze brain responses across tasks without a joint model of the psychological processes. The method boosts statistical power in small studies with specific cognitive focus by analyzing them jointly with large studies that probe less focal mental processes. Our approach improves decoding performance for 80\% of 35 widely-different functional-imaging studies. It finds commonalities across tasks in a data-driven way, via common brain representations that predict mental processes. These are brain networks tuned to psychological manipulations. They outline interpretable and plausible brain structures. The extracted networks have been made available; they can be readily reused in new neuro-imaging studies. We provide a multi-study decoding tool to adapt to new data.

\section*{Author summary}

Brain-imaging findings in cognitive neuroscience often have low statistical power, despite the availability of functional imaging data across hundreds of studies. Yet, with current analytic frameworks, combining data across studies that map responses to different tasks discards the nuances of the cognitive questions they ask. In this paper, we propose a new approach for fMRI analysis, where a predictive model is used to extract the shared information from many studies together, while respecting their original paradigms.
Our method extracts cognitive representations that associate a wide
variety of functions to specific brain structures. This provides quantitative improvements and cognitive insights when analyzing together 35 task-fMRI studies; the breadth of the functional data we consider is much higher than in previous work. Reusing the representations learned by our approach also improves statistical power in studies outside the training corpus.

\section*{Introduction}

Cognitive neuroscience uses functional brain imaging to probe the brain
structures underlying mental processes. The field is accumulating neural
activity responses to specific psychological manipulations. The diversity of
studies that probe different mental processes gives a big picture on cognition
\cite{yarkoni_large-scale_2011}. However, as brain mapping has progressed in
exploring finer aspects of mental processes, the statistical power of studies
has stagnated or even decreased \cite{poldrack_scanning_2017}---although
sample size is increasing over years, it has not kept pace with the reduction
of effect size. As a result, many, if not most individual studies often have
low statistical power \cite{button_power_2013}. Large-scale efforts address this
issue by collecting data from many subjects
\cite{vanessen_human_2012,miller2016multimodal}.
For practical reasons, 
these efforts however focus on a small number of cognitive tasks.
In contrast, establishing a complete view of the links between brain
structures and the mental processes that they implement 
requires 
varied cognitive tasks \cite{poldrack_cognitive_2006}, each
crafted to recruit different mental processes. In this paper, we
develop an analysis methodology that pools data
across many task-fMRI studies to increase both
statistical power and cognitive coverage. 
Standard meta analyses can only address \textit{commonalities}
across studies, as they require casting mental manipulations in a consistent
overarching cognitive theory. They can bring statistical power at the
cost of coverage and specificity in the cognitive processes.
On the opposite, our approach uses the \textit{specific} psychological manipulations of
each study and extracts shared
information from the brain responses across paradigms. As a result, it improves markedly the
statistical power of mapping brain structures to mental processes. We 
demonstrate these benefits on 35 functional-imaging studies, 
all analyzed accordingly to their individual experimental paradigm.

Interpreting overlapping brain responses calls for multivariate
analyses such as brain decoding \cite{haxby2001}.
Brain decoding uses machine learning to predict mental processes from
the observed brain activity. It is a crucial tool to associate functions
to given brain structures. Such inference endeavor calls for decoding
across cognitive paradigms
\cite{poldrack_decoding_2009}. Indeed, a single study does not provide
enough psychological manipulations to characterize well the functions of
the brain structures that it activates \cite{poldrack_cognitive_2006},
while covering a broader set of
cognitive paradigms gives more precise functional descriptions.
Moreover, the statistical power of functional data is limited by the
sample size \cite{button_power_2013}. A single study seldom provides more than 
few hundreds of observations, which is well below machine-learning
standards. Open repositories of brain
functional images \cite{poldrack_toward_2013,gorgolewski2015neurovault}
bring the hope of large-scale decoding with
much larger sample sizes.

Yet, shoehorning such a diversity of studies into
a decoding problem requires daunting manual annotation to build
explicit correspondences across cognitive paradigms.
%
We propose a different approach: we treat the decoding of each study as a
single task in a multi-task linear decoding
model~\cite{ando_framework_2005, xue_multi-task_2007}. The parameters of
this model are partially shared across studies to enable discovering
potential commonalities. Model fitting---the training step of machine
learning---is performed jointly, using non-convex training and regularization techniques~\cite{kingma_adam:_2014,srivastava_dropout:_2014}. We thus learn to perform simultaneous decoding in many studies, to leverage the
brain structures that they implicitly share.
The extracted structures provide universal priors of functional mapping that
improve decoding on new studies and can readily be reused in subsequent
analyzes.

Models that generalize in measurable ways to new cognitive
paradigms would ground broader pictures of cognition
\cite{varoquaux2019predictive}. However, they face
the fundamental roadblock that each
cognitive study frames a particular question and resorts to specific task
oppositions without clear counterpart 
in other studies \cite{newell_you_1973}.
In particular, a cognitive fMRI study results in 
\emph{contrast} brain maps, each of which corresponds
to an elementary psychological manipulation, often unique to a given protocol.
Analyzing contrast maps across studies requires to model the relationships
between protocols, which is a challenging problem. It has been tackled by
labeling common aspects of
psychological manipulations across studies, 
to build decoders that describe aspects of
unseen paradigms \cite{wager_fmri-based_2013, varoquaux_atlases_2018}.
This annotation
strategy is however difficult to scale up to a large set of studies as
it requires expert knowledge on each study. The lack of complete
cognitive ontologies to decompose
psychological manipulations into mental processes
\cite{poldrack_brain_2016}
makes it even harder.

To overcome these obstacles, our multi-study decoding approach relies on the \textit{original}
labels of each study. Instead of relabeling data
into a common ontology, the method extracts data-driven common cognitive dimensions.
Our guiding hypothesis is that activation
maps may be accurately decomposed into latent components that form the neural
building blocks underlying cognitive processes~\cite{barrett_future_2009}.
%
%
This modelling overcomes the
limitations of single-study cognitive subtraction
models \cite{poldrack_brain_2016}. In particular, we show that it
improves statistical power in individual studies:
it gives better decoding performance for a vast majority
of studies, and the improvement is particularly pronounced for studies with a
small number of subjects.
%
Our implicit modelling of functional information has the further advantage of providing explainable predictions. It decomposes
the common aspects of psychological manipulations across studies
onto latent factors, supported by spatial brain networks that are \textit{interpretable}
for neuroscience. These form by themselves a valuable resource for brain
mapping: a functional atlas tuned to
jointly decoding the cognitive information conveyed by various protocols.
%
The trained model is a deep \emph{linear} model. Building a linear model is
important to bridge with classic decoding
techniques in neuroimaging and ensures interpretability of intermediary representations.

\begin{figure}
     \includegraphics[width=\textwidth]{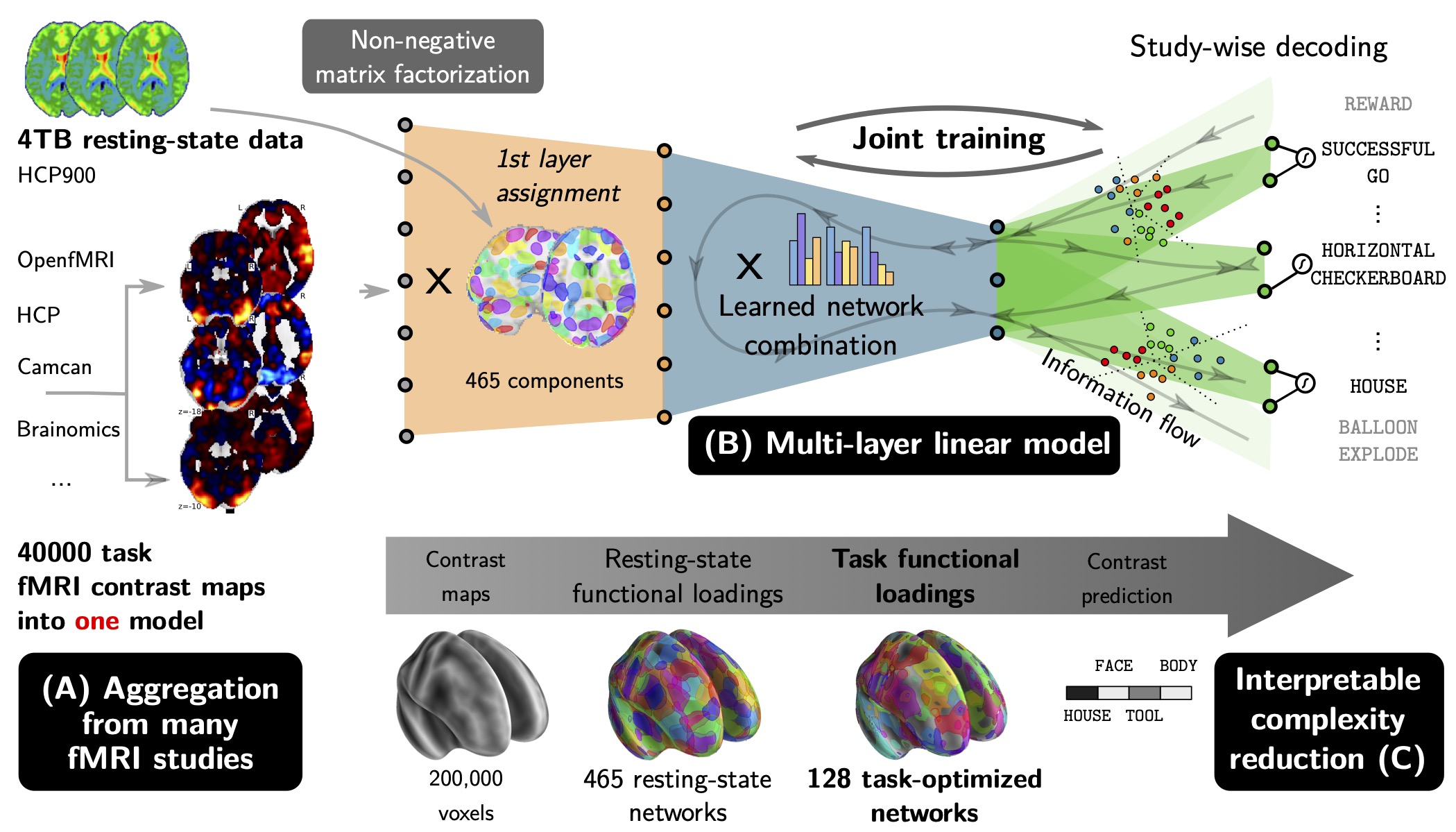}
    \caption{\textbf{General description of our multi-study decoding approach.} We perform inter-subject decoding using a shared three-layer model trained on multiple studies. An initial layer projects the input images from all studies onto functional networks learned on resting-state data. Then, a second layer combines the functional networks loadings into common meaningful cognitive subspaces that are used to perform decoding for each study in a third layer. The second and third layers are trained jointly, fostering transfer learning across studies.}\label{fig:nature_abstract}
  \end{figure}

\section*{Materials and methods}

We first give an informal overview of the contributed methods for multi-study
decoding. We review the mathematical foundations of the methods in a second part---a complete description is provided in \nameref{S1A_appendix}. Finally, we describe how we validate the performance and usability of the approach. A preliminary version of our method was described in \cite{mensch2017learning}, with important differences and a less involved validation (\nameref{S1B_appendix}).

\begin{table*}[!h]
    \caption{\textbf{Training and experiment set of fMRI studies}. Note that even though some tasks are similar, they may feature different contrasts. Task correspondence is not encoded explicitly in our model. Supplementary \autoref{table:all_contrasts} lists each contrast used in each study.}\label{table:studies}
        \centering
\footnotesize
\begin{tabular}{p{9cm}ll}
\toprule
                                                                                                                                   Study and task description & \# contrasts & \# subjects \\
\midrule
                                                                                                       \cite{amalric2016origins} High level math \& Localizer &           31 &          30 \\
                                                                                                                           \cite{pinel2007} The ARCHI project &           30 &          78 \\
                                                                                                               \cite{papadopoulos_brainomics_2017} Brainomics &           19 &          94 \\
                                                                                                                          \cite{shafto_cambridge_2014} CamCAN &            5 &         605 \\
                                                                             \cite{cauvet2012traitement,hara2009neural} Music structure \& Sentence structure &           19 &          35 \\
                                                                                                     \cite{devauchelle2009sentence} Sentence/music complexity &           25 &          20 \\
                                                                                                    \cite{schonberg2012decreasing} Balloon Analog Risk-taking &           12 &          16 \\
                                                                                                 \cite{aron2006long} Baseline trials \& Classication learning &            7 &          17 \\
                                                                                                                          \cite{xue2007neural} Rhyme judgment &            3 &          13 \\
                                                                                                                         \cite{tom_neural_2007} Mixed-gambles &            4 &          16 \\
                                                                                                      \cite{jimura_neural_2014} Plain or mirror-reversed text &            9 &          14 \\
                                                                                                                             \cite{xue2008common} Stop-signal &            6 &          20 \\
                                                                                          \cite{aron2007triangulating} Conditional stop-signal \& Stop-signal &           12 &          13 \\
                                        \cite{cohen2009generality} Balloon analog risk task \& Emotion regulation \& Stop-signal \& Temporal discounting task &           23 &          24 \\
 \cite{foerde2006modulation} Classification probe without feedback \& Dual-task weather classification \& Single-task weather classification \& Tone-counting &           14 &          14 \\
                                                                                                          \cite{ds017} Classification learning \& Stop-signal &           11 &           8 \\
                                                                                                          \cite{ds017} Classification learning \& Stop-signal &           11 &           8 \\
                                                                                                         \cite{alvarez2002} Cross-language repetition priming &           17 &          13 \\
                                                                                                       \cite{poldrack2001interactive} Classification learning &            3 &          13 \\
                                                                                                                                      \cite{ds101} Simon task &            8 &           7 \\
                                                                                                                   \cite{haxby2001} Visual object recognition &           13 &           6 \\
                                                                                                       \cite{duncan2009consistency} Word \& object processing &            6 &          49 \\
                                                                                                                \cite{wager2008prefrontal} Emotion regulation &           26 &          34 \\
                                                                                                                          \cite{moran2012social} False belief &            7 &          36 \\
                                                                                                         \cite{uncapher_dissociable_2011} Incidental encoding &           26 &          18 \\
                               \cite{gorgolewski2013test} Covert verb generation \& Line bisection \& Motor \& Overt verb generation \& Overt word repetition &           11 &          10 \\
                                                                                            \cite{collier_comparison_2014} Auditory oddball \& Visual oddball &            8 &          17 \\
          \cite{gauthier2012temporal} Continuous house vs face \& Discontinuous house (800ms) vs face \& Discoutinuous house (400ms) vs face \& House vs face &           30 &          11 \\
                                                                                        \cite{gauthier2012temporal} Continuous house vs face \& House vs face &           23 &          13 \\
                                                                                                                \cite{barch2013} The Human Connectome Project &           23 &         786 \\
                                                                                                               \cite{henson_parametric_2011} Face recognition &            5 &          16 \\
                                                                                                                      \cite{knops2009} Arithmetic \& Saccades &           26 &          19 \\
                                                                                                       \cite{poldrack_phenome-wide_2016} UCLA LA5C consortium &           24 &         189 \\
                                                                                                    \cite{pinel2013} Foreign language \& Localizer \& Saccade &           34 &          65 \\
                                                                                  \cite{vagharchakian2012temporal} Auditory compression \& Visual compression &           14 &          16 \\
                                                                                                                                                        Total &          545 &        2343 \\
\bottomrule
\end{tabular}
   
\end{table*}

\subsection*{Method overview}

The approach has three main components, summarized in \autoref{fig:nature_abstract}:
aggregating many fMRI studies, training a deep linear model, and reducing this model to extract \textit{interpretable} intermediate
representations. These representations are readily reusable to apply the
methodology to new data.
Building upon the increasing availability of public task-fMRI data, we
gathered statistical maps from many task studies, along with rest-fMRI data from large repositories,
to serve as training data for our predictive model
(\autoref{fig:nature_abstract}A). Statistical maps are obtained by standard
analysis, computing z-statistics maps for either base conditions, or for
contrasts of interest when available. We use 40,000 subject-level
contrast maps from 35 different studies (detailed in \autoref{table:studies}), with 545 different contrasts; a few are
acquired in cohorts of hundreds
of subjects (\eg \acsfont{HCP}, CamCan, \acsfont{LA5C}), but most of them feature more
common sample sizes of 10 to 20 subjects. These studies use different
experimental paradigms, though most recruit related aspects of
cognition (\eg motor, attention, judgement tasks, object recognition).

We use machine-learning classification techniques for inter-subject decoding.
Namely, we associate each brain activity contrast map with a predicted contrast
class, chosen among the contrasts of the map's study. For this, we propose a
linear classification model featuring \textit{three} layers of transformation
(\autoref{fig:nature_abstract}B). This architecture reflects our working
hypothesis: cognition can be represented on basic functions distributed
spatially in the brain. The first layer projects contrast maps onto $k=465$
functional units learned from resting-state data. This first dimension reduction
should be interpreted as a projection of the brain signal onto small, smooth and
connected brain regions, tuned to capture the resting-state brain signal with a
fine grain. The second layer performs dimension reduction and outputs an
embedding of the brain activity into $l=128$ features that are \emph{common} across
studies. The embedded data from each study are then classified into their
respective contrast class using a study-specific classification output from the
third layer, in a setting akin to multi-task learning~(see
\cite{pan_survey_2010} for a review).

The second layer and the third layer are jointly extracted from the task-fMRI
data using regularized stochastic optimization. Namely, the shared brain
representation is optimized simultaneously with the third layer that performs
decoding for every study. In particular, we use dropout regularization
\cite{kingma_variational_2015} in the layered model and stochastic optimization \cite{kingma_adam:_2014} to obtain good out-of-sample performance.

Study-specific decoding is thus performed on a shared
low-dimensional brain representation. This representation is supported
on 128 different combinations of the first 465 functional units identified with
resting-state data. These combinations form diffuse networks of brain regions,
that we call \textit{multi-study task-optimized networks} (\acs{MSTONs}). MSTONs differ from the notion of
brain networks in the neuroscience literature---the later are typically obtained using a low-rank factorization of resting-state data, with a much lower number of components ($k \approx 20$) than what we use to extract the \textit{functional units} of the first layer.

As we will show, projecting data onto MSTONs improves across-subject
predictive accuracy, removing confounds while preserving the cognitive signal. Interpretability is guaranteed by the linearity of the model and a post-training identification of stable directions in the space of latent representations. These networks capture a general multi-study representation of the cognitive signal contained in statistical maps.

\subsection*{Mathematical modelling}

Following this informal description, we now review the mathematical foundations of our decoding approach. The complete
descriptions of the predictive models and of the training algorithms are
provided in \nameref{S1A_appendix}.

We consider $N$ task-fMRI
studies, that we use for functional decoding. In this setting, each study $j$
features $n^j$ subjects, for which we compute $c^j$ different contrasts maps,
using the General Linear Model \cite{friston_statistical_1994}. Masking them using a grey-mask filter in the MNI space, we obtain a
set of $z$-maps $(\x_j^i)_{j \in [1, c^j n^j]}$, in $\RR^p$, that summarizes the
effect on brain activations of the psychological conditions $(y_i^j)_{i \in [1,
c^j n^j]}$. The goal of functional decoding is to learn a predictor from
$z$-maps to psychological conditions, namely a function $f^j : \RR^p \to
[1,c^j]$. This predictor will be evaluated on unseen subjects for validation.

\paragraph{Linear decoding with shared parameters.} In our setting, we couple
the predictors $(f^j)_{j \in [N]}$ by forcing them to share parameters. Each
study corresponds to a classification task, and we cast the problem as
multi-task learning (as first considered in \cite{ando_framework_2005}). For
this, we consider a given $z$-map $\x^j_i$ in study $j$. We compute the
predicted psychological condition using a factorized linear model:
\begin{equation}
    \hat y_i^j = f^j(\x^j_i) = \argmax_{k \in [1, c^j]} {(\U^j \L \D \x^j_i + \b^j)}_k.
\end{equation}
The matrix $\D \in \RR^{k \times p}$ and $\L \in \RR^{l \times k}$ contain the
basis for performing two successive projection of the $z$-map $\x_j^i$ onto
low-dimension spaces. Those parameters are shared over all studies $j \in [N]$
and form the first and second layer of our model. The matrix $\U^j \in \RR^{l
\times c^j}$ and the bias vector $\b^j \in \RR^{c^j}$ are the parameters of a
multi-class linear classification model that labels the projected map $\L \D
\x_i^j$ with a psychological condition within the study $j$. Those parameters
are specific to each study~$j$, and form the third layer of our model.

\paragraph{First layer training from resting-state data.} The first dimension
reduction, contained in the matrix $\D \in \RR^{k \times p}$, is learned using
external resting-state data, from the HCP project \cite{vanessen_human_2012}.
Voxel time-series are stacked in a data matrix $\X \in \RR^{n \times p}$ (with
$4$ millions brain-images), that is factorized so that $\X \approx \D \A$, with
$\D$ non-negative and sparse (i.e. with mostly null coefficients). This forces
the elements of $\D$ to delineate localized functional units. We use a sparse
non-negative matrix factorization objective \cite{mairal_online_2010} and a
recent scalable matrix factorization algorithm \cite{mensch_stochastic_2017} to
learn $\D$, as detailed in \nameref{S1A_appendix}. The non-negativity
constraint allows to interpret functional units as a soft parcellation of the brain. We do not use additional spatial constraints, as non-negative sparse matrix factorization with $k=465$ components readily finds smooth connected regions.

\paragraph{Joint training of the second and third layer.} The matrix $\L$ and
the multiple matrices $(\U^j)_{j \in [n]}$ and intercepts $(\b^j)_j$ are trained
jointly to minimize the objective
\begin{equation}
    \min_{\L, 
    \{\U^j, j \in [N]\}
    } \sum_{j=1}^N \frac{1}{n^j} 
    \sum_{i=1}^{c^j n^j} \ell_j(\U^j \L \D \x_i^j + \b^j, y_i^j),
\end{equation}
where $\ell_j$ is the standard $\ell_2$-regularized multinomial loss function for training a linear
model with $c^j$ classes (see \nameref{S1A_appendix} for
details). This objective is trained using Adam
\cite{kingma_adam:_2014}; at each step, we select a batch of examples from one study. To prevent specialization of the rows of
matrix $\L$ to specific studies, we add a dropout noise
\cite{srivastava_dropout:_2014} to the activations $\D \x^j_i$ and $\L \D \x^j_i$
during training. 

\paragraph{Model consensus.} Although the atoms of $\D$ are naturally
interpretable, the fact that the product $\U^j \L$ can always be rewritten as
$\U^j \M^{-1} \M \L$ for an invertible matrix $\M$ prevents us from directly
identifying meaningful directions in the low-dimensional space spanned by $\L
\D$. On the other hand, we found this space to be remarkably stable across
training runs. We therefore propose an ensemble technique to extract a
non-negative matrix $\bar \L \in \RR^{l \times k}$ such that $\bar \L \D$ captures
meaningful directions (as above-mentioned non-negativity enables us to interpret MSTONs as soft brain parcellations).

For this, we train $R$ decoding models with different sampling order and
initialization, to obtain ${(\L_r)}_{r \in [R]}$. We stack these matrices into a
tall matrix $\tilde \L \in \RR^{l\,R \times k}$, that we factorize as $\tilde \L
= \K \bar \L$, with $\bar \L \in \RR^{l \times k}$ non-negative and sparse. This
is in turn (see 
\nameref{S1A_appendix}) yields a consensus model ${(\D, \bar \L,
(\bar \U^j, \bar \b^j)}_{j \in [N]})$, where $\bar \L \D \in \RR^{l \times p}$
is sparse and non-negative. It therefore holds interpretable brain networks,
learned in a supervised manner from many studies---those form the MSTONs.

\paragraph{Layer widths.} We chose $k=465$ and $l=128$ as those are a good compromise
between model performance and interpretability---trade-offs in choosing the
number of functional units $k$ for fMRI analysis are discussed in e.g.
\cite{dadi2019fine}, and we compare the model performance for different $l$ in 
\autoref{fig:varying_l}, \nameref{S1B_appendix}. Choosing $l$ smaller than the number of classes
enforces a low-rank structure over the set of 545 classification maps.

\subsection*{Validation}

\paragraph{Quantitative measurements.} The benefits of multi-study decoding may vary from study to study, and a single
number cannot properly quantify the impact of our approach. We measure decoding \textit{accuracy}
on left-out subjects (half-split, repeated 20 times) for each study. For each split and each study, we compare
the scores obtained by our model to results obtained by simpler baseline
decoders, that classify contrast maps separately for each study, and directly
from voxels. To analyse the impact of our method on the prediction accuracy
specifically for each contrast, we also report the \textit{balanced-accuracy}
for each predicted class. For completeness, we
report mean accuracy gain and the number of studies for which multi-study
decoding improves accuracy---those hint at the benefit that one may expect when
applying the method to a new fMRI study. Mathematical definitions of the metrics in use are reported in \autoref{sec:metrics}. 

\paragraph{Exploring MSTONs.} Our model optimizes its second and third
layers to project brain images on representations that help decoding.
These representations boil down to MSTONs combinations: MSTONs form a valuable output of the model, as they can easily be reused to project data for new decoding tasks.
We provide 2D and 3D views of the MSTONs, showing how they cover the
brain.
We evaluate the importance of each network for decoding a certain
contrast by computing the cosine similarity between the MSTON and the
classification map associated with this contrast.
We represent these contrasts' names as specified in their original
studies with word-clouds, with a size increasing with their similarity
with a given MSTON.

\paragraph{Classification maps.} As our model is linear, we qualitatively compare the classification maps that it
yields with maps obtained with a baseline single-study voxel-level decoding approach. For both approaches, we compute the
correlation matrix between classification maps to uncover potential clusters of
similar maps, using hierarchical clustering \cite{gower_minimum_1969}. We compare this correlation matrix in term of how clustered it is, using the cophenetic correlation coefficient~\cite{sokal_comparison_1962} and the mean  absolute cosine similarity between maps.

\subsection*{Reusable tools and resources}

Our approach can be used to improve statistical power of decoding in new fMRI
studies. To facilitate its use, we have released resources and the
\textit{cogspaces} library (\url{http://cogspaces.github.io}). We include
software to train the models. Pre-trained MSTONs networks (with associated
word-clouds) can be downloaded and inspected on a dedicated page (\url{https://cogspaces.github.io/assets/MSTON/components.html}).%
The statistical maps used
in the present study may be downloaded using our library, or on \url{neurovault.org}. The published MSTON networks hold the representations extracted from
the 35 studies that we have considered; some of them are shown in
\autoref{fig:nature_latent}.
\section*{Results}

We first detail the quantitative improvements brought by our approach, before exploring these results from a cognitive neuroscience point of view.

\subsection*{Improved statistical performance of multi-study decoding}

\begin{figure}[t]
    \centering
    \hspace{-.4\linewidth}%
   \includegraphics[width=1.4\linewidth]{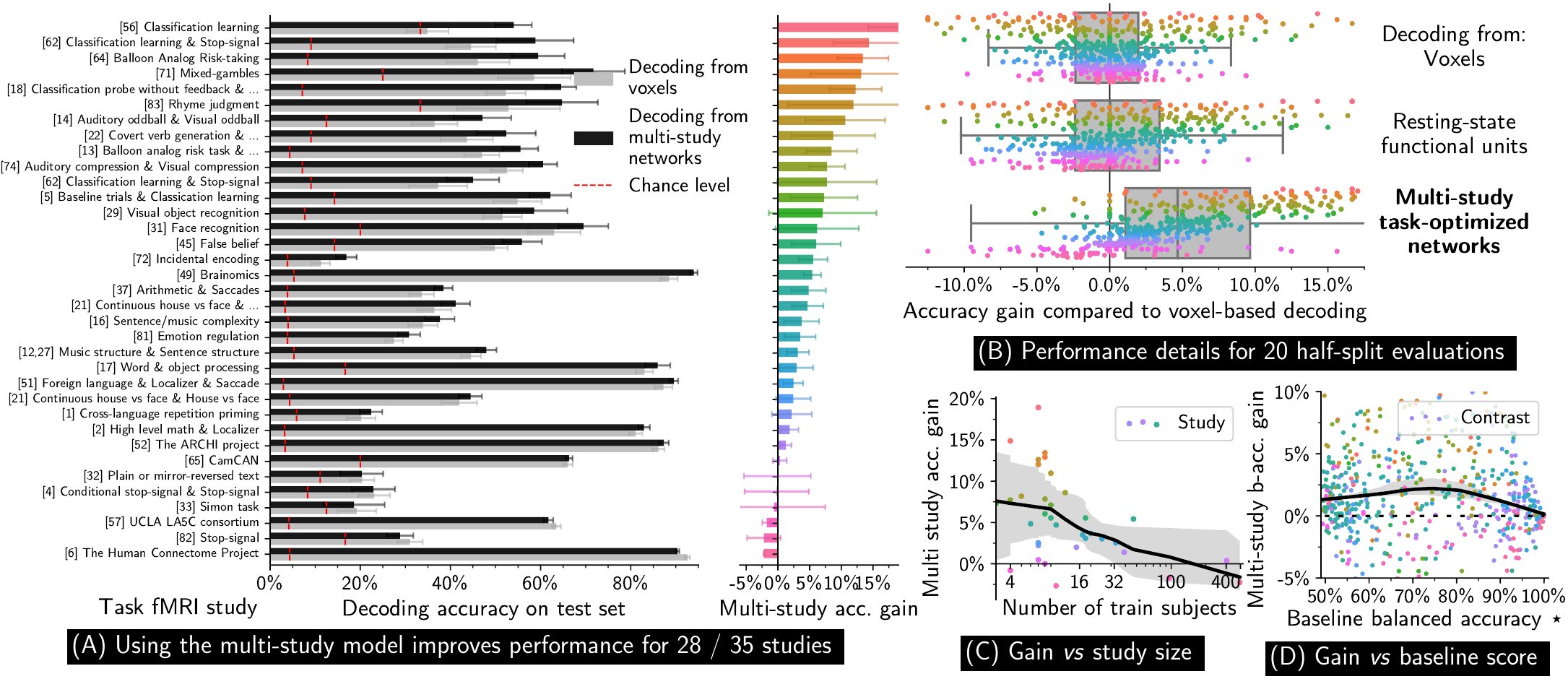}
  \caption{\textbf{Quantitative performance of multi-study
  decoding.} \textbf{(A)} Multi-study decoding improves the performance of  cognitive task prediction across subjects for most studies. \textbf{(B)} Overall, decoding
from task-optimized networks leads to a mean improvement accuracy of
5.8\% compared to voxel or networks based approaches. Each point corresponds to a study and a train/test split.
\textbf{(C)} Studies of typical size strongly benefit from transfer
learning, whereas little information is gained for very large studies.
\textbf{(D)} Contrasts that are moderately difficult to decode benefit
most from transfer.  Error bars are calculated over 20 random
data half-split. $^\star$\textbf{(D)} shows \textit{per-contrast} balanced accuracy (50\% chance level), whereas \textit{per-study} classification accuracy is used everywhere else. Numbers are reported in \autoref{table:accuracy}
}
\label{fig:nature_quantitative}
\end{figure}

\begin{figure}
   \includegraphics[width=\linewidth]{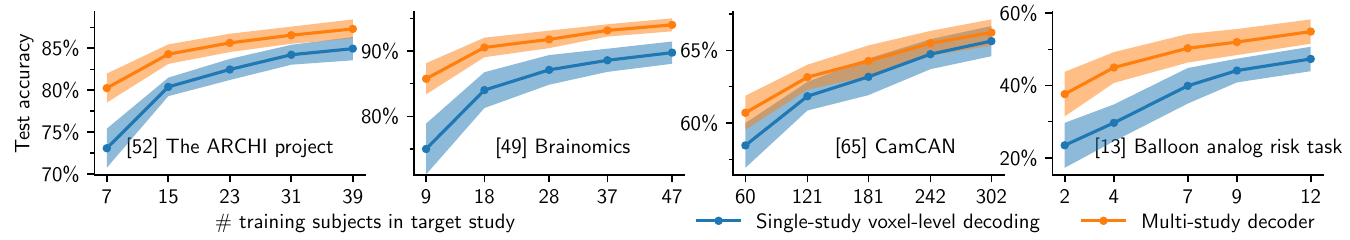}
  \caption{\textbf{Varying accuracy improvement with study size.} Training an MSTON decoder increases decoding accuracy
  for many studies (see \autoref{fig:nature_quantitative}A). Gains are
higher as we reduce the number of training subjects in target
studies---pooling multiple studies is especially useful to decode studies
performed on small cohorts. Error bars are calculated over 20 random data half-splits.}\label{fig:training_curves}
\end{figure}

Decoding from multi-study task optimized networks gives quantitative
improvements in prediction of mental processes, as summarized in
\autoref{fig:nature_quantitative}. For
28 out of the 35 task-\acs{fMRI} studies that we consider, the
\acs{MSTON}-based decoder outperforms single-study decoders
(\autoref{fig:nature_quantitative}A). It improves accuracy by
17\% for the top studies, with a mean gain of 5.8\% (80\% experiments with net increase, 4.8\% median gain) across studies and
cross-validation splits (\autoref{fig:nature_quantitative}B).
 \textit{Jointly} minimizing errors on every study constructs
second-layer representations that are efficient for many study-specific
decoding tasks; the second layer parameters therefore incorporate information
from all studies. This shared representation enables 
information transfer
among the many decoding tasks performed by the third layer---predictive accuracy is thus improved thanks to \textit{transfer learning}. Although we have
not explicitly modeled how mental processes or psychological manipulations 
are related across experiments,
our quantitative results show that these relations can be captured by the
model---encoded into the second layer---to improve decoding performance.

Studies with diverse cognitive focus benefit from using multi-study
modeling. The different decoding tasks have varying difficulties---we report performance sorted by chance level in 
\autoref{fig:sorted_quantitative}, \nameref{S1B_appendix}.
Among the highest accuracy gains, we find cognitive control (stop-signal), classification
studies, and localizer-like protocols. Our corpus contains many of such
studies; as a result, multi-study decoding has access to many more
samples to gather information on the associated cognitive networks.
The activation of these networks is better captured in the shared part of the model,
thereby leading to the observed improvement. In
contrast, for a few studies, among which \acsfont{HCP} and \acsfont{LA5C}, we
observe a slight negative transfer effect. This is not surprising---as
\acsfont{HCP} holds 900 subjects, it may not benefit from the aggregation of
much smaller studies; \acsfont{LA5C} focuses on higher-level cognitive
processes with limited counterparts in the other studies, which precludes
effective transfer.

\autoref{fig:nature_quantitative}B shows that simply projecting data onto
resting state functional networks instead of using our three-layer model does       
not significantly improve decoding, although the net accuracy gain varies from
study to study. Adding a second \textit{task-optimized}---supervised---dimension
reduction is thus necessary to improve overall decoding accuracy. Functional
contrasts that are either easy or very hard to decode do not benefit much from
multi-study modeling, whereas classes with a balanced-accuracy around 80\%
experience the largest decoding improvement
(\autoref{fig:nature_quantitative}). We attribute this to two causes:
easy-to-decode studies do not benefit from the extra signal provided by other
studies, while some studies in our corpus are simply too hard to decode due to
a low signal-to-noise ratio. \autoref{fig:nature_quantitative}D shows that the
benefit of multi-study modeling is higher for smaller studies, confirming that
the proposed method boosts their inter-subject decoding performance.

In \autoref{fig:training_curves}, we vary the number of training subjects in
target studies, and compare the performance of the multi-study decoder with a more standard one.
We observe that the smaller the study size, the larger
the performance gain brought by multi-study modeling. Transfer learning in
inter-subject decoding is thus
particularly effective for small studies (\eg 16 subjects), that still constitute the essential
of task-fMRI studies.
To confirm this effect, we trained a multi-study model on a subset of $15$ subjects per study, considering studies that comprise more than $30$ subjects. In this case, the transfer learning effect is positive for all studies (\autoref{fig:quantitative_subjects} in \nameref{S1B_appendix}),
 including those for which negative transfer was observed when using full cohorts.

Finally, we show in \autoref{fig:comparison_method_transfer} (\nameref{S1B_appendix})
that training a three-layer model and reusing the first two layers as a fixed dimension reduction when decoding a new study improves decoding accuracy on average. The extracted functional networks (MSTONs) thus provide a study-independent prior that is likely to improve decoding for studies probing different cognitive questions than the ones considered in the training corpus.

\begin{figure*}
  \centering
    \hspace{-.4\linewidth}\includegraphics[width=1.4\linewidth]{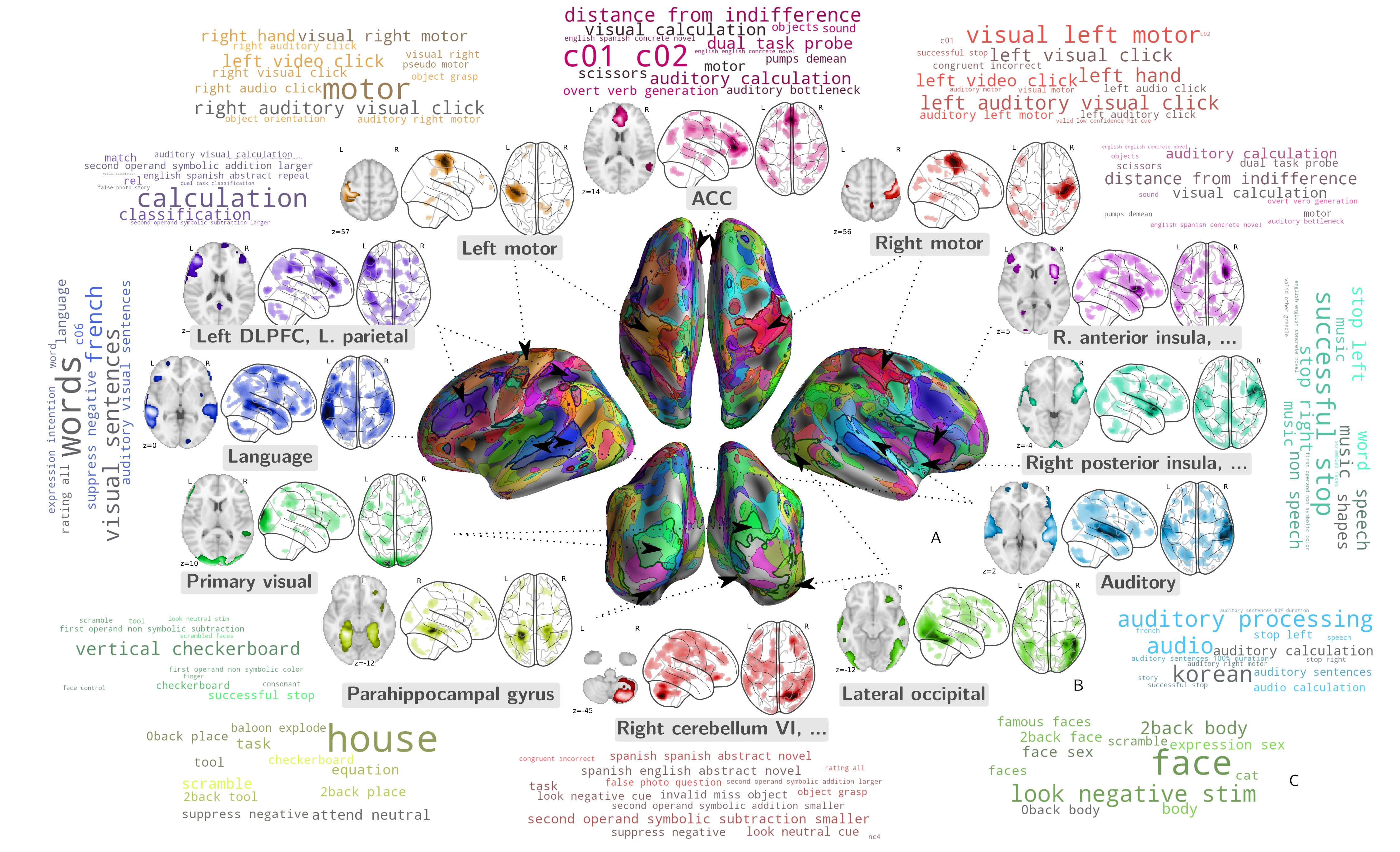}
  \caption{
    \textbf{Visualization of some of task-optimized networks.} Our approach 
  learns networks that are important for decoding across studies. These
networks are individually focal and collectively well spread across the
cortex. They are readily associated with the cognitive tasks that they
contribute to predict. We display a selection of these networks on the
cortical surface \textbf{(A)} and in 2D transparency \textbf{(B)}, named
with the salient anatomical brain region they recruit, along with a
word-cloud~\textbf{(C)} representation of the stimuli whose likelihood
increases with the network activation. The words in this word cloud
are the terms used in the contrast names by the investigators; they
are best interpreted in the context of the corresponding studies.
}\label{fig:nature_latent}
\end{figure*}

\subsection*{Multi-study task-optimized networks capture broad cognitive domains}

We outline the contours of the 128 extracted \acs{MSTONs} in \autoref{fig:nature_latent}A. The
networks almost cover the entire cortex, a consequence of the broad
coverage of cognition of the studies we gathered. Task-optimized networks 
must indeed capture information to predict 545 different
cognitive classes from the resulting distributed brain activity.
Brain regions that are systematically
recruited in task-\acs{fMRI} protocols, \eg motor cortex, auditory
cortex, and primary visual cortex, are finely segmented by \acs{MSTON}: they appear
in several different networks. Capturing information
in these regions is crucial for decoding many contrasts in our corpus,
hence the model dedicates a large part of its representation capability to it.
As decoding requires capturing distributed activations,
\acs{MSTON} are formed of multiple regions (\autoref{fig:nature_latent}B). For instance, both parahippocampal
gyri appear together in the yellow bottom-left network.

\begin{figure}[t]
    \centering
     \includegraphics[width=\linewidth]{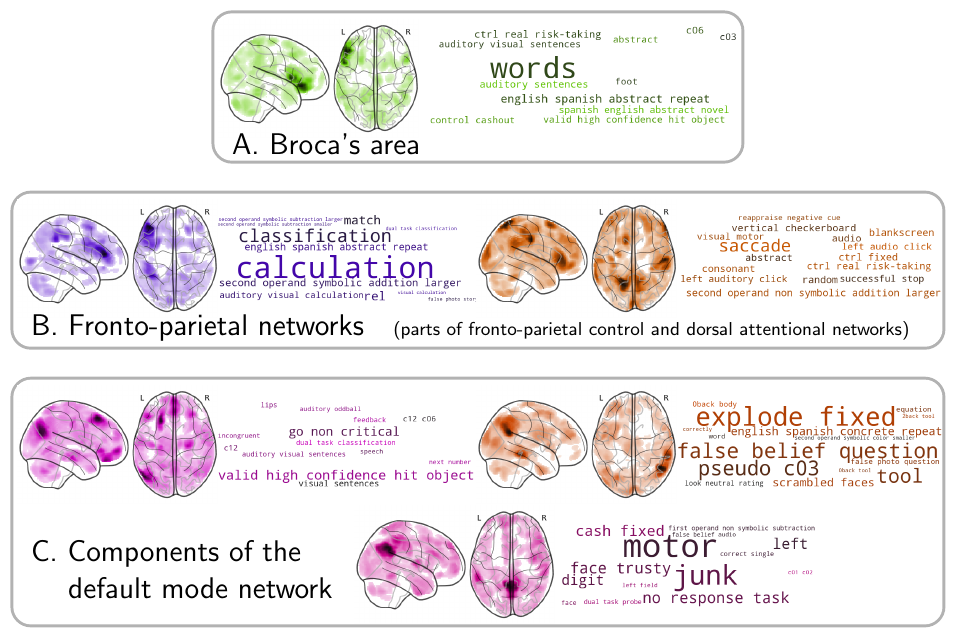}
    \caption{\textbf{Task-optimized networks associated with high-level
functions}. Some MSTONs outline brain-circuits that are associated with
language, e.g. Broca's area \textbf{(A)}, or more abstract functions,
e.g. fronto-parietal networks \textbf{(B)} or even part of 
the default mode network \textbf{(C)}. Those networks are more distributed than the ones displayed in \autoref{fig:nature_latent}, but are associated with relatively interpretable word-clouds.}\label{fig:high_level}
\end{figure}

Most importantly, \autoref{fig:nature_latent}B-C show that the model
relates extracted
\acs{MSTONs} to specific cognitive information.
The MSTONs each play a role in decoding a subset of contrasts.
Components may capture
low-level or high-level cognitive signal, though the low-level components
are easier to interpret.
Indeed, at a lower level, they outline the primary visual cortex,
associated with contrasts such as checkerboard stimuli, and both hand motor cortices,
associated with various tasks demanding motor functions. At a higher
level, some interpretable components single out the left \acs{DLPFC} and the \acs{IPS} in separate
networks,
used to decode tasks related to calculation and comparison. Others
delineate the language network and the right posterior insula,
important in decoding tasks involving music~\cite{hara2009neural}. 
Yet another MSTON delineates Broca's area,
associated with language tasks (\autoref{fig:high_level}).

Inspecting the tasks associated with the MSTONs reveals
structure-function links. Once again, the results are more interpretable
for low-level functions, although some well-known
high-level functional associations are also well captured. For instance,
several components on \autoref{fig:nature_latent}
involve brain regions recruited across a wide variety of
tasks, such as the anterior insula, engaged in
auditory and visual tasks \cite{braver2001anterior}
and considered to tackle ambiguous perceptual information, or
the \acs{ACC},
associated with tasks with affective components
\cite{stevens2011anterior} and
reward-based decision making \cite{bush2002dorsal}. Some MSTONs are more distributed, but correspond to well-known
patterns brain activity. For example, \autoref{fig:high_level} show
components that reveal parts
the default mode networks --associated with baseline conditions,
theory-of-mind tasks and prospection
\cite{raichle2001default,spreng2010patterns}--,
parts of
the fronto-parietal control network --associated with a variety of
problem-solving tasks \cite{gratton2018control}-- and the dorsal attentional network --associated
with visuo-spatial attention tasks such as saccades
\cite{ptak2012frontoparietal}. 

Visualizing MSTONs along with word-clouds serves essentially an illustratory
purpose. It yields more interpretable results with focal networks than with
distributed networks. In both cases, the words in the contrasts related
to the given MSTONs capture documented structure-function associations. Interpretability may be improved
by reducing the number of extracted networks, at the cost of a quantitative loss
in performance. In particular, with $k=128$ components, the default mode network
is split across several MSTONs (\autoref{fig:high_level}). Such a
splitting is common for high-dimensional decomposition of the fMRI
signal, as noted in resting state \cite{kiviniemi2009functional}, as a
network such as the default-mode network has different sub-units with
distinct functional contributions \cite{leech2011fractionating}.
Conversely, some contrast maps are correlated with several distributed MSTONs, as illustrated in \autoref{fig:duplicated-components} (\nameref{S1B_appendix}).

\begin{figure}[t]
    \centering
     \includegraphics[width=\linewidth]{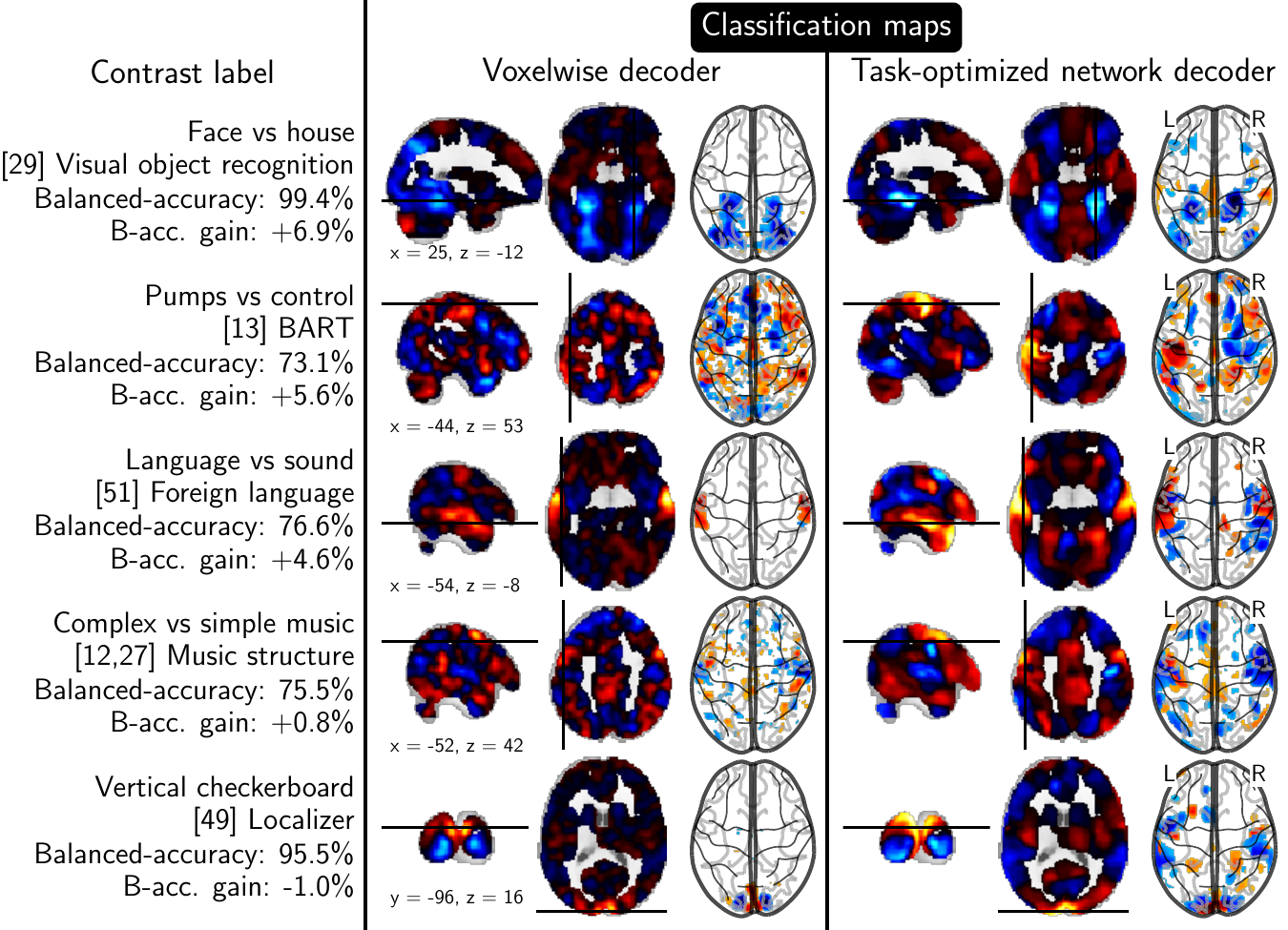}
    \caption{\textbf{Classification maps obtained from multi-study decoding} (right). The maps are smoother and more focused on functional
  modules than when decoding
  from voxels (left). For contrasts for which there is a performance boost (top
  of the figure), relevant brain regions are better delineated, as clearly
  visible on the face vs house visual-recognition opposition, in which the
  fusiform gyrus stands out better. B-acc stands for balanced accuracy using multi-study decoding (see text).}\label{fig:nature_classifs}

\end{figure}

\subsection*{Impact of multi-study modeling on classification maps}

To better understand how multi-study training and layered
representations improve decoding performance, we compare classification maps
obtained using our model to standard decoder maps in
\autoref{fig:nature_classifs}. For contrasts with significant accuracy gains, the classification maps are less noisy and more
focal. They single out determinant regions more clearly, \eg the fusiform face
area (\acs{FFA}, row 1) in classification maps for the face-vs-house contrast, or
the left motor cortex in maps (row 2) predicting pumping action in \acs{BART} tasks~\cite{schonberg2012decreasing}.
The language network is typically better delineated by our model (row 3), and so
is the posterior insula in music-related contrasts (row 4). These improvements
are due to two aspects: First, projecting onto a lower dimension subspace has a
denoising effect on contrast maps, that is already at play when projecting onto
simple resting-state functional networks. Second, multi-study training finds more
scattered classification maps, as these combine complex
MSTONs, learned on a large set of brain images. Our method slightly decreases performance for a
small fraction of contrasts, such as maps associated with vertical checkerboard
(row 5), a condition well localized and easy to decode from the original data.
Our model renders them too much distributed, an unfortunate consequence of
multi-study modeling.

We also compare original input contrast maps to their transformation by the
projection on task-optimized networks (\autoref{fig:nature_classifs_proj} in \nameref{S1B_appendix}
). Projected data are more focal, i.e. spatial
variations that are unlikely to be related to cognition are smoothed. This
offers a new angle on the quantitative results
(\autoref{fig:nature_quantitative}): brain activity expressed as the activation
of these networks captures better cognition and allows decoders to generalize
better across subjects than when classifying raw input directly.

\paragraph{Information transfer among classification maps.}

\begin{figure}
      \centering
    \hspace{-.3\linewidth}%
     \includegraphics[width=1.3\linewidth]{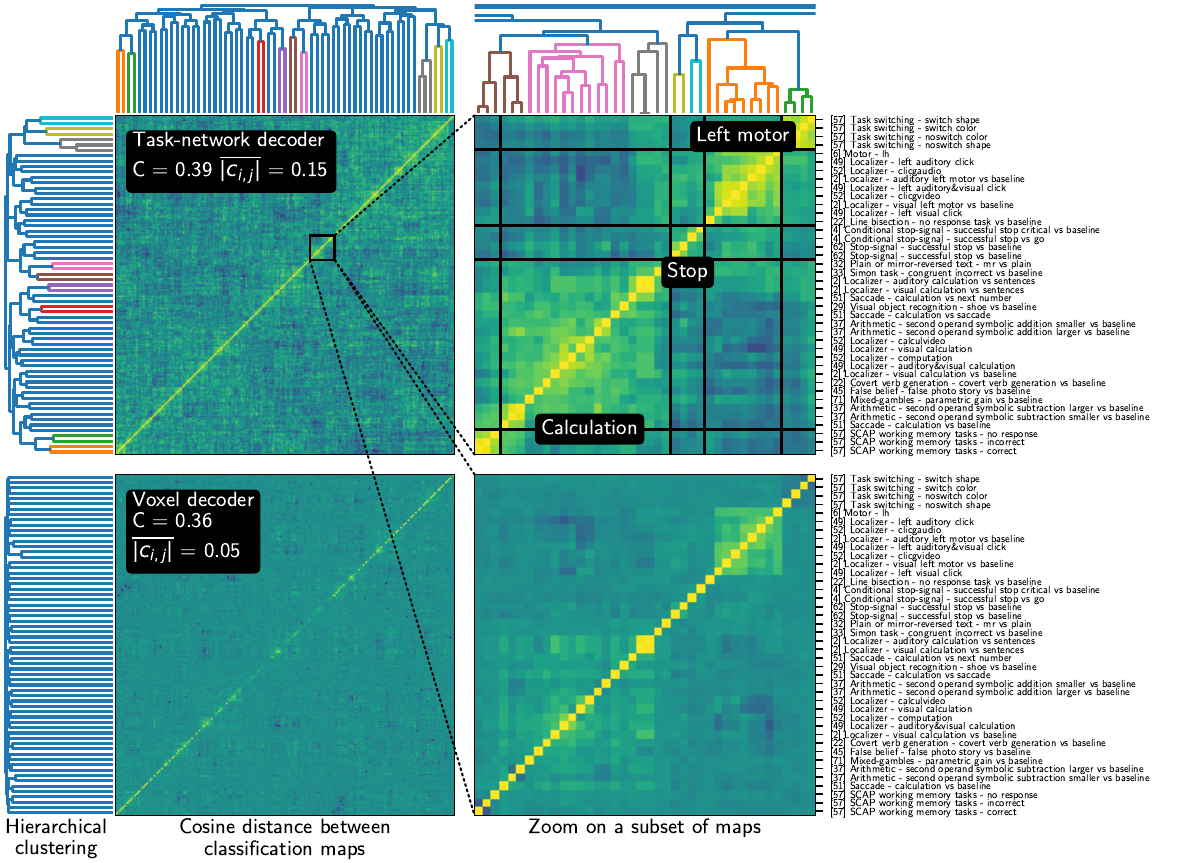}
    \caption{
        \textbf{Cosine similarities between classification maps,} obtained with our multi-study decoder (top) and with decoders learned separately (bottom),
        clustered using average-linkage hierarchical clustering.
        The classification maps obtained when decoding from task-optimized networks are more easily clustered into cognitive-meaningful groups using hierarchical clustering---the cophenetic coefficient of the top clustering is thus higher. Maps may also be compared using the similarities of their loadings on MSTONs, with similar results.
    }\label{fig:nature_dendro}
\end{figure}

In \autoref{fig:nature_dendro}, we compare the correlation between the 545 classification maps obtained using
a multi-study decoder and using simple functional networks decoders. Classification maps learned using task-optimized networks are more correlated on average, and
hierarchical clustering reveals a sharper correlation structure. 
This is because the whole classification matrix is
low-rank (rank $l = 128 < c = 545$) and influenced by the many studies we consider---the classification maps of our model are supported by networks relevant for cognition. As a consequence, it is easier to cluster
maps into meaningful groups using hierarchical clustering based on cosine
distances. For instance, we outline inter-study groups of maps related to
left-motor functions, or calculation tasks. Hierarchical clustering on baseline maps is less successful: the associated dendrogram is less structured, and the distortion introduced by clusters is higher (as suggested by the smaller cophenetic coefficient). Clusters are harder to identify,
due to smaller contrast in the correlation matrix. Multi-study training thus
acts as a regularizer, by forcing correlation across maps with discovered relations. This regularization partly
explains the increase in decoding accuracy. 

\section*{Discussion}


The methodology presented in this work harnesses the power of
deep representations to build multi-study decoding models for
brain functional images.
It brings an immediate benefit to functional brain imaging by 
providing a universal way to
improve the accuracy of decoding in a newly acquired dataset.
Decoding is a central tool to draw inferences on which brain
structures implement the neural support of the observed behavior.
It is most often applied to 
task-\acs{fMRI} studies with 30 or less subjects, which tend to lack
statistical power \cite{varoquaux_cross-validation_2018}.
In this regime, aggregating existing studies to a new one using a
multi-study model as the one we propose is likely to improve
decoding performance. This is further evidenced in \autoref{fig:comparison_method_transfer} (\nameref{S1B_appendix}):
 using MSTONs as a decoding basis on a new decoding task outperforms using resting-state networks.
 Of course, such improvement can only occur if the cognitive functions probed by the new study are
 related to the ones probed in the multi-study corpus. We foresee limited benefits when analyzing 
strongly original task fMRI experiments, and experiments studying very
specific and high-level cognitive functions, 
that MSTONs are only partially able to capture (\autoref{fig:high_level}).

With increasing availability of shared and normalized data,
multi-study modeling is an important improvement over
simple decoders, provided that it can adapt to the diversity of cognitive
paradigms. Our \emph{transfer-learning} model has such
flexibility, as it does not require explicit correspondence across
experiments. Beyond quantitative benefits --the gain in
prediction accuracy-- the models also brings
qualitative benefits, facilitating the interpretation of decoding maps
(\autoref{fig:nature_classifs}).
Pooling subjects across studies effectively increases the sample size, as
advocated by \cite{poldrack_scanning_2017}. The resulting increase in
statistical power for cognitive modeling will help addressing the
reproducibility challenge outlined by \cite{button_power_2013}. In
our setting, each study (or \textit{site}) provides a single decoding objective,
which is predicting one contrast among all other contrasts from this study. This is a validated approach in decoding \cite{bzdok_semisupervised_2015}. As some
studies use different fMRI tasks, we may also use one decoding
objective per \textit{task}, with similar quantitative improvement in performance (see \autoref{fig:per-task-quantitative} in \nameref{S1B_appendix}).

Our modeling choices were driven by the recent successes of deep non-linear models in
computer vision and medical imaging. However, we
were not able to increase performance by departing from linear models:
introducing non linearities in our models provides no improvement on left-out accuracy. On the other hand, we have shown that
pooling many \acs{fMRI}
data sources enables to learn deeper models, although these
remain linear.
Techniques developed in deep
learning prove useful to fit models that generalize well
across subjects: using dropout regularization~\cite{srivastava_dropout:_2014} and
advanced stochastic gradient techniques~\cite{kingma_adam:_2014} 
is crucial for successful
transfer and good generalization performance.

Sticking to linear models brings the benefit of
easy interpretation of decoding models. The use of sparsity and non-negativity in the training and consensus phase
allow to obtain interpretable networks. Using sparsity only in each phase (as
originally advocated by \cite{varoquaux_multisubject_2011}) yields
\enquote{contrast} networks with both positive and negative regions, that are
harder to interpret (see also \cite{dadi2019fine}). In particular,
this limits the occurence of non-zero weights that reflect noise
suppression \cite{haufe_interpretation_2014}.

The models capture information relevant for many decoding tasks in their
internal representations.
From these internals, we extract interpretable cognitive networks, inspired by matrix factorization techniques
used to interpret computer vision 
models~\cite{olah_building_2018}. The good predictive performance of MSTONs networks
(\autoref{fig:nature_quantitative} and \autoref{fig:comparison_method_transfer} in \nameref{S1B_appendix}) provides quantitative support for
their decomposition of brain function. Extracting a universal basis of
cognition is beyond the scope of a single \acs{fMRI} study, and should be
done by analysis across many studies. We show that, across studies, a joint
predictive model finds meaningful approximations of
atomic cognitive functions spanning a wide variety of mental processes
(\autoref{fig:nature_latent}). This methodology provides a step forward
towards defining mental processes in a quantitative manner, which remains a
fundamental challenge in psychology~\cite{uttal_new_2001,poldrack_toward_2013}. 
Yet, in the present work, the delineation of atomic cognitive functions
remains coarse and incomplete. This is likely
due to the limited scope of our corpus, and to the fact that we automatically align the cognitive functions
 probed by the various studies of the corpus. Expert annotation of mental
process involved in the studies could greatly help establishing a clearer
picture.

Our approach differs from commonly-used decomposition techniques in fMRI
analysis (e.g. ICA \cite{mckeown_analysis_1998}, or dictionary learning
\cite{varoquaux_multisubject_2011}), that are used to extract \textit{functional
networks}. These techniques optimize an unsupervised reconstruction objective over
resting-state data, in effect capturing co-occurrence of brain activity
across distributed locations.
They have traditionally been used with few components (e.g. $k \approx 20$). In
contrast, after the first decomposition, performed without
information from the tasks,
we extract the MSTONs components to optimize the decoding performance
on many tasks. Leaving a
systematic comparison between MSTONs and classical functional networks for future work, we already make two observations. 
First, a fraction of functional networks extracted by unsupervised
methods
support non-Gaussian noise patterns
in the BOLD time-series, and permits noise suppression
\cite{perlbarg2007corsica,salimi2014automatic}. Typically, only a
fraction of the networks extracted in an ICA analysis is interpreted.
MSTONs, on the other
hand, optimize a supervised objective and focus on the fraction of the
BOLD signal related to the tasks.
Second, MSTONs (despite being more noisy) appears more skewed
towards known coordinated brain networks (Figs \ref{fig:nature_latent} and \ref{fig:high_level}),
 that differs from the networks recruited at rest (see e.g. \cite{laird_jocn} 
 for a comparison of task and rest brain networks).

We use many different
 fMRI studies to distill MSTONs across various tasks.
This data aggregation approach requires little supervison. The flip side
is that it leads to coarse results by nature: our approach is obviously
 not sufficient to recover the detailed brain-to-mind mapping,
collective knowledge
of psychologists and neuroscientists,
that has emerged from decades of research on multimodal datasets and
careful behavioral experiments. Specific brain-to-mind associations
are best resolved with dedicated
experiments using experimental-pyschology paradigms tailored to the
question at hand. Other data than fMRI, for instance more invasive, may
also provide stronger evidence. For instance a double dissociation in
brain-lesion patients give unambiguous evidence of distinct cognitive
processes via distant neural supports, as with Broca and Wernicke's
separation of language understanding and generation
\cite{friederici1998first}, or the more recent
teasing out of emotional
and cognitive empathy \cite{shamay2009two}.

Finally, the current version of our framework does not model 
explicit inter-subject variability, and is rather focused on extracting
commonalities across subjects. Future work may augment multi-study decoding
with such information, as obtained by e.g., hyperalignment
techniques~\cite{haxby_common_2011}.

\section*{Conclusion}

The success of using distributed representations to bridge cognitive tasks
supports a system-level view on 
how brain activity supports cognition.



Our multi-study model will become increasingly useful to brain imaging as
the number of available studies grows.
Such a growth is driven by the
steady increase of publicly shared brain-imaging data, facilitated by online
neuroimaging platforms and increased
standardization~\cite{gorgolewski_brain_2016,poldrack_scanning_2017}. 
With a larger corpus of studies, the proposed methodology 
has the potential to build even better universal
priors that overall improve statistical power for functional brain
imaging.
As such, multi-study decoding provides a path towards knowledge
consolidation in functional neuroimaging and cognitive neuroscience.



\bibliography{article}

\begin{thebibliography}{100}

\bibitem{yarkoni_large-scale_2011}
Yarkoni T, Poldrack RA, Nichols TE, Van~Essen DC, Wager TD.
\newblock Large-Scale Automated Synthesis of Human Functional Neuroimaging
  Data.
\newblock Nature Methods. 2011;8(8):665--670.

\bibitem{poldrack_scanning_2017}
Poldrack RA, Baker CI, Durnez J, Gorgolewski KJ, Matthews PM, Munaf{ò} MR,
  et~al.
\newblock Scanning the Horizon: Towards Transparent and Reproducible
  Neuroimaging Research.
\newblock Nature Reviews Neuroscience. 2017;18(2):115--126.

\bibitem{button_power_2013}
Button KS, Ioannidis JPA, Mokrysz C, Nosek BA, Flint J, Robinson ESJ, et~al.
\newblock Power Failure: Why Small Sample Size Undermines the Reliability of
  Neuroscience.
\newblock Nature Reviews Neuroscience. 2013;14(5):365--376.

\bibitem{vanessen_human_2012}
Van~Essen DC, Ugurbil K, Auerbach E, Barch D, Behrens TEJ, Bucholz R, et~al.
\newblock The {Human} {Connectome} {Project}: A Data Acquisition Perspective.
\newblock {N}euro{I}mage. 2012;62(4):2222--2231.

\bibitem{miller2016multimodal}
Miller KL, Alfaro-Almagro F, Bangerter NK, Thomas DL, Yacoub E, Xu J, et~al.
\newblock Multimodal Population Brain Imaging in the UK Biobank Prospective
  Epidemiological Study.
\newblock Nature Neuroscience. 2016;19(11):1523--1536.

\bibitem{poldrack_cognitive_2006}
Poldrack RA.
\newblock Can cognitive processes be inferred from neuroimaging data?
\newblock Trends in cognitive sciences. 2006;10(2):59--63.

\bibitem{haxby2001}
Haxby JV, Gobbini IM, Furey ML, Ishai A, Schouten JL, Pietrini P.
\newblock Distributed and Overlapping Representations of Faces and Objects in
  Ventral Temporal Cortex.
\newblock Science. 2001;293(5539):2425--2430.

\bibitem{poldrack_decoding_2009}
Poldrack RA, Halchenko YO, Hanson SJ.
\newblock Decoding the Large-Scale Structure of Brain Function by Classifying
  Mental States Across Individuals.
\newblock Psychological Science. 2009;20(11):1364--1372.

\bibitem{poldrack_toward_2013}
Poldrack RA, Barch DM, Mitchell J, Wager TD, Wagner AD, Devlin JT, et~al.
\newblock Toward Open Sharing of Task-Based {{fMRI}} Data: The {{OpenfMRI}}
  Project.
\newblock Frontiers in Neuroinformatics. 2013;7:12.

\bibitem{gorgolewski2015neurovault}
Gorgolewski KJ, Varoquaux G, Rivera G, Schwarz Y, Ghosh SS, Maumet C, et~al.
\newblock NeuroVault.org: A Web-Based Repository for Collecting and Sharing
  Unthresholded Statistical Maps of the Human Brain.
\newblock Frontiers in Neuroinformatics. 2015;9:8.

\bibitem{ando_framework_2005}
Ando RK, Zhang T.
\newblock A Framework for Learning Predictive Structures from Multiple Tasks
  and Unlabeled Data.
\newblock Journal of Machine Learning Research. 2005;6:1817--1853.

\bibitem{xue_multi-task_2007}
Xue Y, Liao X, Carin L, Krishnapuram B.
\newblock Multi-Task Learning for Classification with Dirichlet Process Priors.
\newblock Journal of Machine Learning Research. 2007;8(Jan):35--63.

\bibitem{kingma_adam:_2014}
Kingma DP, Ba J.
\newblock Adam: A Method for Stochastic Optimization.
\newblock In: International Conference for Learning Representations; 2015.

\bibitem{srivastava_dropout:_2014}
Srivastava N, Hinton GE, Krizhevsky A, Sutskever I, Salakhutdinov R.
\newblock Dropout: A Simple Way to Prevent Neural Networks from Overfitting.
\newblock Journal of Machine Learning Research. 2014;15(1):1929--1958.

\bibitem{varoquaux2019predictive}
Varoquaux G, Poldrack RA.
\newblock Predictive models avoid excessive reductionism in cognitive
  neuroimaging.
\newblock Current opinion in neurobiology. 2019;55:1--6.

\bibitem{newell_you_1973}
Newell A.
\newblock You Can't Play 20 Questions with Nature and Win: {{Projective}}
  Comments on the Papers of This Symposium.
\newblock Visual Information Processing. 1973; p. 1--26.

\bibitem{wager_fmri-based_2013}
Wager TD, Atlas LY, Lindquist MA, Roy M, Woo CW, Kross E.
\newblock An {fMRI}-Based Neurologic Signature of Physical Pain.
\newblock New England Journal of Medicine. 2013;368(15):1388--1397.

\bibitem{varoquaux_atlases_2018}
Varoquaux G, Schwartz Y, Poldrack RA, Gauthier B, Bzdok D, Poline JB, et~al.
\newblock Atlases of cognition with large-scale human brain mapping.
\newblock PLoS computational biology. 2018;14(11):e1006565.

\bibitem{poldrack_brain_2016}
Poldrack RA, Yarkoni T.
\newblock From Brain Maps to Cognitive Ontologies: Informatics and the Search
  for Mental Structure.
\newblock Annual Review of Psychology. 2016;67(1):587--612.

\bibitem{barrett_future_2009}
Barrett LF.
\newblock The Future of Psychology: Connecting Mind to Brain.
\newblock Perspectives on Psychological Science. 2009;4(4):326--339.

\bibitem{mensch2017learning}
Mensch A, Mairal J, Bzdok D, Thirion B, Varoquaux G.
\newblock Learning Neural Representations of Human Cognition Across Many {fMRI}
  Studies.
\newblock In: Advances in Neural Information Processing Systems; 2017. p.
  5883--5893.

\bibitem{amalric2016origins}
Amalric M, Dehaene S.
\newblock Origins of the Brain Networks for Advanced Mathematics in Expert
  Mathematicians.
\newblock Proceedings of the National Academy of Sciences.
  2016;113(18):4909--4917.

\bibitem{pinel2007}
Pinel P, Thirion B, Meriaux S, Jobert A, Serres J, Bihan DL, et~al.
\newblock Fast Reproducible Identification and Large-Scale Databasing of
  Individual Functional Cognitive Networks.
\newblock BMC neuroscience. 2007;8:91.

\bibitem{papadopoulos_brainomics_2017}
Papadopoulos~Orfanos D, Michel V, Schwartz Y, Pinel P, Moreno A, Le~Bihan D,
  et~al.
\newblock The {{Brainomics}}/{{Localizer}} Database.
\newblock {N}euro{I}mage. 2017;144:309--314.

\bibitem{shafto_cambridge_2014}
Shafto MA, Tyler LK, Dixon M, Taylor JR, Rowe JB, Cusack R, et~al.
\newblock The {{Cambridge Centre}} for {{Ageing}} and {{Neuroscience}}
  ({{Cam}}-{{CAN}}) Study Protocol: A Cross-Sectional, Lifespan,
  Multidisciplinary Examination of Healthy Cognitive Ageing.
\newblock {BMC Neurology}. 2014;14:204.

\bibitem{cauvet2012traitement}
Cauvet E.
\newblock Traitement des structures syntaxiques dans le langage et dans la
  musique [PhD thesis].
\newblock Paris 6; 2012.

\bibitem{hara2009neural}
Hara N, Cauvet E, Devauchelle AD, Dehaene S, Pallier C, et~al.
\newblock Neural Correlates of Constituent Structure in Language and Music.
\newblock {N}euro{I}mage. 2009;47:S143.

\bibitem{devauchelle2009sentence}
Devauchelle AD, Oppenheim C, Rizzi L, Dehaene S, Pallier C.
\newblock Sentence Syntax and Content in the Human Temporal Lobe: An {fMRI}
  Adaptation Study in Auditory and Visual Modalities.
\newblock Journal of Cognitive Neuroscience. 2009;21(5):1000--1012.

\bibitem{schonberg2012decreasing}
Schonberg T, Fox C, Mumford JA, Congdon C, Trepel C, Poldrack RA.
\newblock Decreasing Ventromedial Prefrontal Cortex Activity During Sequential
  Risk-Taking: An f{MRI} Investigation of the Balloon Analog Risk Task.
\newblock Frontiers in Neuroscience. 2012;6:80.

\bibitem{aron2006long}
Aron AR, Gluck M, Poldrack RA.
\newblock Long-Term Test--Retest Reliability of Functional {MRI} in a
  Classification Learning Task.
\newblock {N}euro{I}mage. 2006;29:1000--1006.

\bibitem{xue2007neural}
Xue G, Poldrack RA.
\newblock The Neural Substrates of Visual Perceptual Learning of Words:
  Implications for the Visual Word Form Area Hypothesis.
\newblock Journal of Cognitive Neuroscience. 2007;19:1643--1655.

\bibitem{tom_neural_2007}
Tom SM, Fox CR, Trepel C, Poldrack RA.
\newblock The Neural Basis of Loss Aversion in Decision-Making Under Risk.
\newblock Science. 2007;315(5811):515--518.

\bibitem{jimura_neural_2014}
Jimura K, Cazalis F, Stover ERS, Poldrack RA.
\newblock The Neural Basis of Task Switching Changes with Skill Acquisition.
\newblock Frontiers in Human Neuroscience. 2014;8.

\bibitem{xue2008common}
Xue G, Aron AR, Poldrack RA.
\newblock Common Neural Substrates for Inhibition of Spoken and Manual
  Responses.
\newblock Cerebral Cortex. 2008;18:1923--1932.

\bibitem{aron2007triangulating}
Aron AR, Behrens TE, Smith S, Frank MJ, Poldrack RA.
\newblock Triangulating a Cognitive Control Network Using Diffusion-Weighted
  Magnetic Resonance Imaging {(MRI)} and Functional {MRI}.
\newblock The Journal of Neuroscience. 2007;27:3743--3752.

\bibitem{cohen2009generality}
Cohen JR.
\newblock The Development and Generality of Self-Control [PhD thesis].
\newblock University of the City of Los Angeles; 2009.

\bibitem{foerde2006modulation}
Foerde K, Knowlton B, Poldrack RA.
\newblock Modulation of Competing Memory Systems by Distraction.
\newblock Proceedings of the National Academy of Science.
  2006;103:11778--11783.

\bibitem{ds017}
Rizk-Jackson A, Aron AR, Poldrack RA. Classification Learning and Stop-Signal
  (one Year Test-Retest); 2011.
\newblock \url{https://openfmri.org/dataset/ds000017}.

\bibitem{alvarez2002}
Alvarez RP, Jasdzewski G, Poldrack RA.
\newblock Building Memories in Two Languages: An {fMRI} Study of Episodic
  Encoding in Bilinguals.
\newblock In: Society for Neuroscience Abstracts; 2002. p. 179.12.

\bibitem{poldrack2001interactive}
Poldrack RA, Clark J, Pare-Blagoev E, Shohamy D, Creso~Moyano J, Myers C,
  et~al.
\newblock Interactive Memory Systems in the Human Brain.
\newblock Nature. 2001;414(6863):546--550.

\bibitem{ds101}
Kelly A, Milham M. Simon Task; 2011.
\newblock \url{https://openfmri.org/dataset/ds000101}.

\bibitem{duncan2009consistency}
Duncan K, Pattamadilok C, Knierim I, Devlin J.
\newblock Consistency and Variability in Functional Localisers.
\newblock {N}euro{I}mage. 2009;46:1018--1026.

\bibitem{wager2008prefrontal}
Wager TD, Davidson ML, Hughes BL, Lindquist MA, Ochsner KN.
\newblock Prefrontal-Subcortical Pathways Mediating Successful Emotion
  Regulation.
\newblock Neuron. 2008;59:1037--1050.

\bibitem{moran2012social}
Moran JM, Jolly E, Mitchell JP.
\newblock Social-Cognitive Deficits in Normal Aging.
\newblock The Journal of Neuroscience. 2012;32:5553--5561.

\bibitem{uncapher_dissociable_2011}
Uncapher MR, Hutchinson JB, Wagner AD.
\newblock Dissociable Effects of Top-Down and Bottom-Up Attention During
  Episodic Encoding.
\newblock The Journal of Neuroscience: The Official Journal of the Society for
  Neuroscience. 2011;31(35):12613--12628.

\bibitem{gorgolewski2013test}
Gorgolewski KJ, Storkey A, Bastin ME, Whittle IR, Wardlaw JM, Pernet CR.
\newblock A Test-Retest {fMRI} Dataset for Motor, Language and Spatial
  Attention Functions.
\newblock GigaScience. 2013;2(1):6.

\bibitem{collier_comparison_2014}
Collier AK, Wolf DH, Valdez JN, Turetsky BI, Elliott MA, Gur RE, et~al.
\newblock Comparison of Auditory and Visual Oddball {fMRI} in Schizophrenia.
\newblock Schizophrenia research. 2014;158:183--188.

\bibitem{gauthier2012temporal}
Gauthier B, Eger E, Hesselmann G, Giraud AL, Kleinschmidt A.
\newblock Temporal Tuning Properties Along the Human Ventral Visual Stream.
\newblock The Journal of Neuroscience. 2012;32:14433--14441.

\bibitem{barch2013}
Barch DM, Burgess GC, Harms MP, Petersen SE, Schlaggar BL, Corbetta M, et~al.
\newblock Function in the Human Connectome: Task-{fMRI} and Individual
  Differences in Behavior.
\newblock {N}euro{I}mage. 2013;80:169--189.

\bibitem{henson_parametric_2011}
Henson RN, Wakeman DG, Litvak V, Friston KJ.
\newblock A Parametric Empirical Bayesian Framework for the {EEG}/{MEG} Inverse
  Problem: Generative Models for Multi-Subject and Multi-Modal Integration.
\newblock Frontiers in Human Neuroscience. 2011;5.

\bibitem{knops2009}
Knops A, Thirion B, Hubbard EM, Michel V, Dehaene S.
\newblock Recruitment of an Area Involved in Eye Movements During Mental
  Arithmetic.
\newblock Science. 2009;324:1583--1585.

\bibitem{poldrack_phenome-wide_2016}
Poldrack RA, Congdon E, Triplett W, Gorgolewski KJ, Karlsgodt K, Mumford JA,
  et~al.
\newblock A Phenome-Wide Examination of Neural and Cognitive Function.
\newblock Scientific Data. 2016;3:160110.

\bibitem{pinel2013}
Pinel P, Dehaene S.
\newblock Genetic and Environmental Contributions to Brain Activation During
  Calculation.
\newblock {N}euro{I}mage. 2013;81:306--316.

\bibitem{vagharchakian2012temporal}
Vagharchakian L, Dehaene-Lambertz G, Pallier C, Dehaene S.
\newblock A Temporal Bottleneck in the Language Comprehension Network.
\newblock The Journal of Neuroscience. 2012;32:9089--9102.

\bibitem{pan_survey_2010}
Pan SJ, Yang Q.
\newblock A {Survey} on {Transfer} {Learning}.
\newblock IEEE Transactions on Knowledge and Data Engineering.
  2010;22(10):1345--1359.

\bibitem{kingma_variational_2015}
Kingma DP, Salimans T, Welling M.
\newblock Variational {Dropout} and the Local Reparameterization Trick.
\newblock In: Advances in Neural Information Processing Systems; 2015. p.
  2575--2583.

\bibitem{friston_statistical_1994}
Friston KJ, Holmes AP, Worsley KJ, Poline JP, Frith CD, Frackowiak RS.
\newblock Statistical Parametric Maps in Functional Imaging: A General Linear
  Approach.
\newblock Human brain mapping. 1994;2(4):189--210.

\bibitem{mairal_online_2010}
Mairal J, Bach F, Ponce J, Sapiro G.
\newblock Online Learning for Matrix Factorization and Sparse Coding.
\newblock Journal of Machine Learning Research. 2010;11:19--60.

\bibitem{mensch_stochastic_2017}
Mensch A, Mairal J, Thirion B, Varoquaux G.
\newblock {Stochastic} {Subsampling} for Factorizing Huge Matrices.
\newblock IEEE Transactions on Signal Processing. 2018;66(1):113--128.

\bibitem{dadi2019fine}
Dadi K, Varoquaux G, Machlouzarides-Shalit A, Gorgolewski KJ, Wassermann D,
  Thirion B, et~al.
\newblock Fine-grain atlases of functional modes for fMRI analysis.
\newblock To appear in NeuroImage. 2020;.

\bibitem{gower_minimum_1969}
Gower JC, Ross GJ.
\newblock Minimum Spanning Trees and Single Linkage Cluster Analysis.
\newblock Journal of the Royal Statistical Society: Series C (Applied
  Statistics);18(1):54--64.

\bibitem{sokal_comparison_1962}
Sokal RR, Rohlf FJ.
\newblock The Comparison of Dendrograms by Objective Methods.
\newblock Taxon; p. 33--40.

\bibitem{braver2001anterior}
Braver TS, Barch DM, Gray JR, Molfese DL, Snyder A.
\newblock Anterior cingulate cortex and response conflict: effects of
  frequency, inhibition and errors.
\newblock Cerebral cortex. 2001;11:825.

\bibitem{stevens2011anterior}
Stevens FL, Hurley RA, Taber KH.
\newblock Anterior cingulate cortex: unique role in cognition and emotion.
\newblock The Journal of neuropsychiatry and clinical neurosciences.
  2011;23(2):121.

\bibitem{bush2002dorsal}
Bush G, Vogt BA, Holmes J, Dale AM, Greve D, Jenike MA, et~al.
\newblock Dorsal anterior cingulate cortex: a role in reward-based decision
  making.
\newblock Proceedings of the National Academy of Sciences. 2002;99:523--528.

\bibitem{raichle2001default}
Raichle ME, MacLeod AM, Snyder AZ, Powers WJ, Gusnard DA, Shulman GL.
\newblock A default mode of brain function.
\newblock Proceedings of the National Academy of Sciences. 2001;98(2):676--682.

\bibitem{spreng2010patterns}
Spreng RN, Grady CL.
\newblock Patterns of brain activity supporting autobiographical memory,
  prospection, and theory of mind, and their relationship to the default mode
  network.
\newblock Journal of cognitive neuroscience. 2010;22:1112.

\bibitem{gratton2018control}
Gratton C, Sun H, Petersen SE.
\newblock Control networks and hubs.
\newblock Psychophysiology. 2018;55:e13032.

\bibitem{ptak2012frontoparietal}
Ptak R.
\newblock The frontoparietal attention network of the human brain: action,
  saliency, and a priority map of the environment.
\newblock The Neuroscientist. 2012;18(5):502--515.

\bibitem{kiviniemi2009functional}
Kiviniemi V, Starck T, Remes J, Long X, Nikkinen J, Haapea M, et~al.
\newblock Functional segmentation of the brain cortex using high model order
  group PICA.
\newblock Human brain mapping. 2009;30(12):3865--3886.

\bibitem{leech2011fractionating}
Leech R, Kamourieh S, Beckmann CF, Sharp DJ.
\newblock Fractionating the default mode network: distinct contributions of the
  ventral and dorsal posterior cingulate cortex to cognitive control.
\newblock Journal of Neuroscience. 2011;31(9):3217--3224.

\bibitem{varoquaux_cross-validation_2018}
Varoquaux G.
\newblock Cross-Validation Failure: {{Small}} Sample Sizes Lead to Large Error
  Bars.
\newblock NeuroImage. 2018;180:68--77.

\bibitem{bzdok_semisupervised_2015}
Bzdok D, Eickenberg M, Grisel O, Thirion B, Varoquaux G.
\newblock Semi-Supervised Factored Logistic Regression for High-Dimensional
  Neuroimaging Data.
\newblock In: Advances in Neural Information Processing Systems; 2015. p.
  3348--3356.

\bibitem{varoquaux_multisubject_2011}
Varoquaux G, {Gramfort} A, {Pedregosa} F, {Michel} V, {Thirion} B.
\newblock Multi-Subject Dictionary Learning to Segment an Atlas of Brain
  Spontaneous Activity.
\newblock Proceedings of the International Conference on Information Processing
  in Medical Imaging. 2011;22:562.

\bibitem{haufe_interpretation_2014}
Haufe S, Meinecke F, Görgen K, Dähne S, Haynes JD, Blankertz B, et~al.
\newblock On the Interpretation of Weight Vectors of Linear Models in
  Multivariate Neuroimaging.
\newblock NeuroImage;87:96--110.

\bibitem{olah_building_2018}
Olah C, Satyanarayan A, Johnson I, Carter S, Schubert L, Ye K, et~al.
\newblock The {{Building Blocks}} of {{Interpretability}}.
\newblock Distill. 2018;3(3):e10.

\bibitem{uttal_new_2001}
Uttal WR.
\newblock The New Phrenology: The Limits of Localizing Cognitive Processes in
  the Brain.
\newblock The MIT press; 2001.

\bibitem{mckeown_analysis_1998}
McKeown MJ, Makeig S, Brown GG, Jung TP, Kindermann SS, Bell AJ, et~al.
\newblock Analysis of {fMRI} Data by Blind Separation Into Independent Spatial
  Components.
\newblock Human Brain Mapping. 1998;6(3):160--188.

\bibitem{perlbarg2007corsica}
Perlbarg V, Bellec P, Anton JL, P{\'e}l{\'e}grini-Issac M, Doyon J, Benali H.
\newblock {CORSICA}: correction of structured noise in {fMRI} by automatic
  identification of {ICA} components.
\newblock Magnetic resonance imaging. 2007;25(1):35--46.

\bibitem{salimi2014automatic}
Salimi-Khorshidi G, Douaud G, Beckmann CF, Glasser MF, Griffanti L, Smith SM.
\newblock Automatic denoising of functional {MRI} data: combining independent
  component analysis and hierarchical fusion of classifiers.
\newblock Neuroimage. 2014;90:449--468.

\bibitem{laird_jocn}
Laird AR, Fox PM, Eickhoff SB, Turner JA, Ray KL, McKay DR, et~al.
\newblock Behavioral interpretations of intrinsic connectivity networks.
\newblock Journal of cognitive neuroscience. 2011;23(12):4022--4037.

\bibitem{friederici1998first}
Friederici AD, Hahne A, Von~Cramon DY.
\newblock First-pass versus second-pass parsing processes in a Wernicke's and a
  Broca's aphasic: electrophysiological evidence for a double dissociation.
\newblock Brain and language. 1998;62:311.

\bibitem{shamay2009two}
Shamay-Tsoory SG, Aharon-Peretz J, Perry D.
\newblock Two systems for empathy: a double dissociation between emotional and
  cognitive empathy in inferior frontal gyrus versus ventromedial prefrontal
  lesions.
\newblock Brain. 2009;132:617.

\bibitem{haxby_common_2011}
Haxby JV, Guntupalli JS, Connolly AC, Halchenko YO, Conroy BR, Gobbini MI,
  et~al.
\newblock A {{Common}}, {{High}}-{{Dimensional Model}} of the
  {{Representational Space}} in {{Human Ventral Temporal Cortex}}.
\newblock Neuron. 2011;72(2):404--416.

\bibitem{gorgolewski_brain_2016}
Gorgolewski KJ, Auer T, Calhoun VD, Craddock RC, Das S, Duff EP, et~al.
\newblock The Brain Imaging Data Structure, a Format for Organizing and
  Describing Outputs of Neuroimaging Experiments.
\newblock Scientific Data. 2016;3:sdata201644.

\bibitem{smith_correspondence_2009}
Smith SM, Fox PT, Miller KL, Glahn DC, Fox PM, Mackay CE, et~al.
\newblock Correspondence of the Brain's Functional Architecture During
  Activation and Rest.
\newblock Proceedings of the National Academy of Sciences.
  2009;106(31):13040--13045.

\bibitem{yeo_organization_2011}
Yeo TBT, Krienen FM, Sepulcre J, Sabuncu MR, Lashkari D, Hollinshead M, et~al.
\newblock The Organization of the Human Cerebral Cortex Estimated by Intrinsic
  Functional Connectivity.
\newblock Journal of Neurophysiology. 2011;106(3):1125--1165.

\bibitem{nocedal_updating_1980}
Nocedal J.
\newblock Updating Quasi-{{Newton}} Matrices with Limited Storage.
\newblock Mathematics of Computation. 1980;35(151):773--782.

\bibitem{neyshabur_implicit_2017}
Neyshabur B.
\newblock Implicit {Regularization} in {Deep} {Learning} [PhD thesis].
\newblock Toyota Technological Institute at Chicago; 2017.

\bibitem{molchanov_variational_2017}
Molchanov D, Ashukha A, Vetrov D.
\newblock Variational {Dropout} {Sparsifies} {Deep} {Neural} {Networks}.
\newblock In: Proceedings of the International Conference on Machine Learning;
  2017. p. 2498--2507.

\bibitem{ioffe_batch_2015}
Ioffe S, Szegedy C.
\newblock {B}atch {N}ormalization: Accelerating Deep Network Training by
  Reducing Internal Covariate Shift.
\newblock In: Proceedings of the International Conference on Machine Learning;
  2015. p. 448--456.

\bibitem{srebro_maximum-margin_2004}
Srebro N, Rennie J, Jaakkola TS.
\newblock Maximum-Margin Matrix Factorization.
\newblock In: Advances in Neural Information Processing Systems; 2004. p.
  1329--1336.

\bibitem{beck_fast_2009}
Beck A, Teboulle M.
\newblock A Fast Iterative Shrinkage-Thresholding Algorithm for Linear Inverse
  Problems.
\newblock SIAM Journal on Imaging Sciences. 2009;2(1):183--202.

\bibitem{rennie_fast_2005}
Rennie JDM, Srebro N.
\newblock Fast Maximum Margin Matrix Factorization for Collaborative
  Prediction.
\newblock In: Proceedings of the International Conference on Machine Learning;
  2005. p. 713--719.

\bibitem{bell_lessons_2007}
Bell RM, Koren Y.
\newblock Lessons from the {{Netflix}} Prize Challenge.
\newblock ACM SIGKDD Explorations Newsletter. 2007;9(2):75--79.

\bibitem{wager_dropout_2013}
Wager S, Wang S, Liang PS.
\newblock Dropout Training as Adaptive Regularization.
\newblock In: Advances in Neural Information Processing Systems; 2013. p.
  351--359.

\bibitem{breiman_bagging_1996}
Breiman L.
\newblock Bagging Predictors.
\newblock Machine Learning. 1996;24(2):123--140.

\bibitem{abraham_machine_2014}
Abraham A, Pedregosa F, Eickenberg M, Gervais P, Mueller A, Kossaifi J, et~al.
\newblock Machine Learning for Neuroimaging with {Scikit-Learn}.
\newblock Frontiers in Neuroinformatics. 2014;8:14.

\bibitem{pedregosa_scikit-learn:_2011}
Pedregosa F, Varoquaux G, Gramfort A, Michel V, Thirion B, Grisel O, et~al.
\newblock Scikit-Learn: Machine Learning in {Python}.
\newblock Journal of Machine Learning Research. 2011;12:2825--2830.

\bibitem{paszke2017pytorch}
Paszke A, Gross S, Chintala S, Chanan G. {P}y{T}orch: Tensors and Dynamic
  Neural Networks in {Python} with Strong {GPU} Acceleration; 2017.

\end{thebibliography}
    
    \section*{Supporting information}

    \paragraph*{S1A Appendix.}\label{S1A_appendix}\textbf{Detailed methods.}
    This appendix discusses technical details of the multi-study decoding approach:
    the specific architecture, a 3-layer linear model, and the deep-learning
    technique used to regularize and train it.
    
    \paragraph*{S1B Appendix.}\label{S1B_appendix}\textbf{Discussion on the model
    design.} In this appendix, we perform supportive experiments to explain the
    observed results, An ablation study of the various model components is provided
    to further support modelling choices.
    
    \paragraph*{S1C Appendix.}\label{S1C_appendix}\textbf{Reproduction details and tables.}
    In this appendix, we provide implementation details for reproducibility, along with tables with quantitative results per contrast.
    

\pagebreak

\appendix

\setcounter{figure}{0}
\renewcommand{\thefigure}{\Alph{figure}}
\setcounter{table}{0}
\renewcommand{\thetable}{\Alph{table}}

\renewcommand{\thesection}{\Alph{section}}

\section*{S1 Appendix}\label{app}

The appendix is structured as follow: in the first section, we formalize the learning setting and method, after describing decoding baselines. In the second section, we perform supportive experiments to explain the observed results, and discuss various alternatives for the model, to further support modelling choices. Finally, we provide reproduction details, along with data and software notes. A visualization of all MSTONs components (that reproduces \url{https://cogspaces.github.io/assets/MSTON/components.html}) is provided for completeness in S1 Components.

\paragraph*{Notations.}We denote scalars, vectors and matrices using lower-case, bold lower-case and
bold upper-case letters, e.g., $x$, $\x$ and $\X$. We denote the
elements of $\X$ by $x_{i,j}$ and its rows by $\x_i$. We write $x^j$ a value that is specific to study number $j$.
 We denote $\bar x$ a value built from an ensemble of value ${(x_s)}_s$. Finally, we write~$[l]$ the set of integers ranging from $1$ to $l$.

\section{Methods}\label{sec:nature_detailed_methods}

We describe in mathematical terms the multi-layer decoder at the center
of our method and provide supporting experiments. We start by formalizing the joint objective loss and the model training process.

\subsection{Inter-subject decoding setting}

We consider $N$ task functional MRI studies (detailed in \autoref{table:studies}), on which we perform inter-subject
decoding. In study number $j$,
$n^j$ subjects are made to perform one (or sometimes several) tasks.
Acquired \acs{BOLD}
time-series are registered to a common template using non-linear spatial registration,
after motion and slice-timing corrections. \acs{BOLD} time-series are then fed
to a standard analysis pipeline, which fits a linear model relating the design matrix of
each experiment to the signal in every voxel. We use the \textit{nistats} 
library for this purpose. From the obtained beta
maps, we compute z-statistics maps, either associated with each of the base
conditions (stimulus or task) of the experiments, or with contrasts defined by
the study's authors. In both cases, z-maps are labeled with a number $1 \leq y
\leq c^j$ that corresponds to $k$-th contrast/base condition (called contrast
in the following). Overall, this produces a set of z-maps ${(\x^j_i)}_{i \in
[c^j n^j]}$ living in~$\RR^p$, where $p$ is the number of voxels, associated with
a sequence of contrast ${(k_i^j)}_{i \in [c^j n^j]}$. The transformation from 3D brain images to 1D vectors is done using a grey-matter mask after alignment with the MNI template. We compare using a grey-matter mask with using a full brain mask in \autoref{sec:grey_matter}.
Inter-subject decoding proposes a model $f_\theta^j : \RR^p \to [1,
  c^j]$ that predicts contrast identity from z-maps, \ie $\hat y_i^j
\triangleq f^j_\theta(\x_i^j)$, where $\theta$ is learned from
training data, and the performance of the model is assessed on
left-out subjects.

\subsection{Baseline voxel-space decoder}

Baseline decoders are linear classifier models defined separately for each study $j$, which take full brain images as input. For every input
map $\x_i$ in $\RR^p$, we compute the logits $\l_i$ in $\RR^{c}$ as
\begin{equation}
\l_i(\W, \b) \triangleq \W \x_i + \b,
\end{equation}
where $\W \in \RR^{c \times p}$ and $\b \in \RR^{c}$ are the parameters of the linear model to be learned for study $j$---we drop the superscript $j$ in this paragraph and the next for simplicity. Logits are transformed into a classification probability vector using the softmax operator.
At test time, we predict the label corresponding to the maximal logit, \ie $\hat y_i = \argmax_{1 \leq y \leq c} l_{i, y}$.
The model is trained on the data $(\x_i, y_i)_{i \in [n]}$ by minimizing the $\ell_2^2$ regularized multinomial classification problem
\begin{align}
    \min_{\substack{\W \in \RR^{c \times p} \\\b \in \RR^{c}}} &- \frac{1}{n} \sum_{i=1}^n 
    \Big(l_{i, y_i}(\W, \b) + \log(\sum_{k=1}^c \exp l_{i, k}(\W, \b)) \Big)
    \\
    &+ \lambda {\Vert \W \Vert}_F^2, \label{eq:classif_loss}
\end{align}
where ${\Vert \cdot \Vert}_F^2$ is the Frobenius norm, that computes ${\Vert \W \Vert}_F^2 \triangleq \sum_{i,j=1}^{c,p} w_{i,j}^2$.

\subsection{Baseline dimension reduced decoder}\label{sec:reduced_multi}

A variant of the voxel-based decoders is obtained by introducing a first-layer dimension reduction learned from resting-state data. This amounts to computing
\begin{equation}
    \l_i(\V, \b, \D) \triangleq \V \D \x_i + \b,
\end{equation}
where $\V$ in $\RR^{c \times k}$ forms the classifying weights of the model, and the matrix $\D$ in $\RR^{k \times p}$ is
\textit{assigned} during training to functional networks learned on
resting-state data, as detailed in \ref{sec:nature_rs}. Multiplying input data by $\D$ projects statistical images onto meaningful
resting-state components, in an attempt to improve classification performance
and reduce computation cost, akin to the methods proposed by~\cite{smith_correspondence_2009, yeo_organization_2011}. The model is trained by solving the convex objective~\eqref{eq:classif_loss} separately for each study, replacing $\W$ by $\V$ in $\RR^{c \times k}$:
\begin{align}\label{eq:classif_loss_red}
    \min_{\substack{\V \in \RR^{c \times k} \\\b \in \RR^{c}}} &- \frac{1}{n} \sum_{i=1}^n 
    \Big(l_{i, y_i}(\V, \b, \D) \\
    &+ \log(\sum_{k=1}^c \exp l_{i, k}(\V, \b, \D)) \Big) + \lambda {\Vert \V \Vert}_F^2.
\end{align}

Our results~(\autoref{fig:nature_quantitative}C) show that decoding from
functional networks is not significantly better than decoding from voxels
directly. For both baselines, the parameter $\lambda$ is found by half-split
cross-valida\-tion. Training is performed using a \acs{L-BFGS} solver~\cite{nocedal_updating_1980}. We use non standardized maps
${(\x_i)}_i$ as input as we observed that standardization hinders performance.

\subsection{Three-layer model description}\label{sec:3layer}

Our three-layer model adds a second shared linear layer in between the
projection on functional networks and the classification models. We still have
\begin{equation}
    \l_i^j(\W^j, \b^j) \triangleq \W^j \x^j_i + \b^j,
\end{equation}
for every z-map $i$ and study $j$. However, we introduce a
coupling between the various parameters $(\W^j)_{j \in [N]}$ of each study:
they should decompose on common basis $\L \D$, where $\L$ is estimated from
the whole corpus of data, and $\D$ is the resting-state dictionary presented above. Formally, we assume that there exist a matrix $\L$ in
$\RR^{l \times k}$ with $l < k < p$, and a set of matrices $(\U^j)_{j \in [N]}$ so that for all $j
\in[N]$, the classification weights of~\eqref{eq:classif_loss} writes
\begin{equation}\label{eq:shared_param}
    \W^j \triangleq \U^j \L \D,\quad\text{where}\quad \U^j \in \RR^{c^j \times l}.
\end{equation}
The matrix $\D$ corresponds to the first-layer weights pictured in
\autoref{fig:nature_abstract}, $\L$ to the second-layer weights, and $(\U^j, \b^j)_j$ to
the various classification heads of the third layer. In this work, we choose $k
= 465$ and $l = 128$. While $\D$ remains fixed, the second-layer matrix $\L$
and the $N$ classification heads $(\U^j)_{j \in [N]}$ are jointly learned
during training, a necessary step toward improving decoding accuracy. The
``shared-layer'' parameterization~\eqref{eq:shared_param} is a common approach
in multi-task learning~\cite{ando_framework_2005, xue_multi-task_2007}, and
should allow \textit{transfer learning} between decoding tasks, under certain
conditions. In our setting, both the data distribution from the different
studies and the classification task associated with each study differ---this is
a particular case of \textit{inductive transfer learning}\footnote{This case is less studied than the classical multi-task
setting where input data are single-source but learning tasks are multiple.}, described by
\cite{pan_survey_2010}.

\paragraph{Modeling.} Without refinement nor regularization, we seek a local minimizer of the
following non-convex objective function, which combines the classification
objectives~\eqref{eq:classif_loss} from all studies, with parameter sharing:
\begin{align}\label{eq:nature_joint}
    \min_{\substack{\L \in \RR^{l \times k}\\
    (\U^j, \b^j)_{j}}} &- \sum_{j=1}^N
    \frac{(n^j)^\beta}{n^j} \sum_{i=1}^{n^j}    
    \Big(l^j_{i, y_i}(\U^j, \b^j, \L) \\
    &- \log(\sum_{k=1}^{c^j} \exp l^j_{i, k}(\U^j, \b^j, \L)) \Big),
\end{align}%
where the dependence on $\D$ is left implicit. The scalar $\beta$ in $[0, 1]$ is
a parameter that regulates the importance of each study in the joint objective,
that we further discuss in \ref{sec:study_weight}. We note that the importance
of the study $j$ to find the latent parameter $\L$ depends on the amplitude of
the gradient $\frac{\partial \ell_j}{\partial \L}$ that does not depend on the
number of tasks $c^j$: in particular, for each study $j$, contrast $1 \leq k
\leq c^j$ and subject $1 \leq i \leq n^j$,  the susceptibility of the loss to
the logits $l_{i,k}^j$ is such that $\frac{\partial \ell_j}{\partial l_{i,k}^j}
\in [-1, 1]$, independent from $c^j$.

\paragraph{Regularization.} We observe that minimizing~\eqref{eq:nature_joint} leads to strong overfitting and low
performance on left-out data, with performance similar to
fitting~\eqref{eq:classif_loss} without regularization, separately for each
study. Adding $\ell_2$ regularization to the second and third layer weights
gives little benefit, as we discuss in \autoref{sec:penalty_transfer}. On the other
hand, introducing \textit{dropout}~\cite{srivastava_dropout:_2014} during
training alleviates the overfitting issue and fosters transfer learning. Dropout is a stochastic regularization method that
prevents the weights from each layer to co-adapt by perturbating them with multiplicative noise during training. It ensures that the
information is well spread across coefficients rows and
columns~\cite{neyshabur_implicit_2017}. In our case, this favors transfer
learning, as it ensures that no single row of $\L$, or in plain words no
task-optimized network, becomes dedicated to a \textit{single} study.
We further compare the different methods that we can use to foster transfer of information
between studies in \autoref{sec:transfer_reg}.

We use the variational flavor of dropout~\cite{kingma_variational_2015} to
make the dropout rate for every study adaptive. This slightly improves
performance compared to binary dropout: every decoding task requires a
different level of regularization, depending on the size of the study and the
hardness of the task, and it is beneficial to estimate it from data. In details, during
training, at every iteration, for every input sample $i$ of a mini-batch from
study $j$, we randomly draw two multiplicative noise matrices
\begin{equation}
    \M_D = \diag([m_{D,t}]_{t \in [k]}), \quad \M_L^j = \diag([m_{L,t}]_{t \in [l]}),
\end{equation}
where $m_{D, t} \sim \mathcal{N}(1, \alpha)$ and $m_{L, t} \sim \mathcal{N}(1,
\alpha^j)$, with $\alpha$ fixed and $\alpha^j$ estimated from
data.\footnote{This \textit{Gaussian} dropout has a similar behavior to the
more commonly used binary dropout with parameter $p = \frac{\alpha}{\alpha +
1}$.} We then compute the noisy logits
\begin{equation}
    \l_i^j \triangleq \U^j \M_L^j \L \M_D \D \x^j_i + \b^j,    
\end{equation}
and use these to compute the loss \eqref{eq:partial_loss}, to which we add a
regularization term that regulates the learning of $\alpha^j$, introduced by
\cite{molchanov_variational_2017}. We compute the gradient with respect to $\L$, $\U^j$,
$\b^j$ using the local reparametrization trick~\cite{kingma_variational_2015}.
We refer to \cite{molchanov_variational_2017} for more details on variational
dropout and a Bayesian grounding of this approach.

\paragraph{Optimization.} We solve the problem~\eqref{eq:nature_joint}
using stochastic optimization. Namely, at each iteration, we compute an
unbiased estimate of the objective~\eqref{eq:nature_joint} and its gradient
with respect to the model parameters, in order to perform a stochastic gradient step. For this, we
randomly choose the study $j$ with a probability proportional to $(n^j)^\beta$, and
consider a mini-batch of z-maps $(\x_i^j)_{j \in B}$ that we use to compute
the unbiased objective estimate
 \begin{equation}\label{eq:partial_loss}
    - \frac{1}{B}
    \sum_{i=1}^n - \Big(l^j_{i, k_i} \log(\sum_{k=1}^c \exp l^j_{i, k}) \Big),
\end{equation}
from which we compute gradients with respect to $\L$, $\U^j$ and $\b^j$.

Optimization is performed using \textit{Adam} \cite{kingma_adam:_2014}, a
flavor of stochastic gradient descent that depends less on the step-size. We
use batch normalization~\cite{ioffe_batch_2015} between the second and third
layer, as it slightly improves performance---it reduces potential negative
transfer learning---and training speed.

\subsection{Resting-state data}\label{sec:nature_rs}

As mentioned above, we use resting-state data to compute the first-layer
weights $\D$ in $\RR^{k \times p}$, where $k = 512$. Such high-order dictionaries are known to perform well for decoding \cite{dadi2019fine}. We consider data from the
\acsfont{HCP900} release, and stack all records to obtain a data matrix $\X$ in
$\RR^{n \times p}$. We then use an online solver \cite{mensch_stochastic_2017} to solve the sparse
non-negative matrix factorization problem
\begin{equation}\label{eq:nature_dl}
    \A, \D \triangleq \argmin_{\D \in \cC, \A \in \RR^{k \times n}} \Vert \X - \A \D \Vert_F^2 + \lambda {\Vert \A \Vert}_F^2,
\end{equation}
where the constraint $\cC = \big\{\D \in \RR^{k \times p}, \D \geq 0, {\Vert
\d_j \Vert}_1\leq 1\:\forall\,j \in [k] \big\}$ enforces every dictionary
component to live in the simplex of $\RR^p$, ensuring sparsity and
non-negativity of the functional networks. The sparsity level is chosen so that
the functional networks $\D$ cover the whole brain with as little overlap as
possible. Larger overlap leads to more correlated activations input to the
second layer, yielding a harder learning problem. With lower coverage,
we would
miss important information to decode some of the predicted psychological conditions.
We refer to \cite{dadi2019fine} for further discussion on selecting sparsity
level when using dictionary learning in fMRI analysis.

\paragraph{Second-layer initialization.} To initialize the weights of the
second layer, we learn a smaller dictionary $\D_l$ in $\RR^{l \times p}$
as in~\eqref{eq:nature_dl}, where $l = 128$. We then compute the initial
weights $\L_l$ so that $\D_l \approx \L_l \D$ using least-square regression.
This way, applying the first two layers initially amount to 
projecting data onto $l = 128$ larger functional networks $\D_l$, which is a reasonable prior
for reducing the dimension of brain statistical maps. Using this resting-state based initialization slightly improves
performance, as we discuss in \autoref{sec:initialization}.

\paragraph{Grey matter restriction.}To help interpreting the obtained mo\-del, we
found it helpful to remove from $\D$ the fraction (9\%) of the functional
networks components located in the white matter and the cerebrospinal fluid
areas, turning $k = 512$ into $k = 465$. We discuss the effect of this
restriction in \autoref{sec:grey_matter}.

\subsection{Model introspection with ensembling}\label{sec:posthoc}

Given any invertible matrix $\M$ in  $\RR^{l \times l}$, the non regularized
version of the objective~\eqref{eq:nature_joint} is left invariant when
transforming $\L$ into $\M \L$ and each $\U^j$ into $\U^j \M^{-1}$. This
prevents us from interpreting the coefficients of $\L$ at the end of the
training procedure, and to retrieve relevant networks by reading the weights of the second weight. The only aspect of $\L$ that remains unchanged after a
linear parameter transformation is its span. Dropout regularization, which
favors the canonical directions in matrix space~\cite{srivastava_dropout:_2014}, should break this symmetry,
but does not help to uncover meaningful directions in the span of $\L$ in
practice.

On the other hand, we found that this span was remarkably stable
across runs on the same data, whether when varying initialization or simply the
order in which data are streamed during stochastic gradient descent. More precisely, we trained our model 100 times with different seeds, and concatenated the weights $(\L_r)_r$ of the second-layer into a big matrix $\bar \L$. We performed a \acs{SVD} on this matrix, and observed that the first $l = 128$ components captured 98\% of the variance of $\bar \L$ when using the same initialization but different streaming order, and 96\% when also using a different random initialization. Despite the many local minima that objective~\eqref{eq:nature_joint} admits, the span of $\L$ thus remains close to some reference span that we can extract with a matrix factorization method.

The above remark suggested the following ensemble method. We run the learning algorithm $R
= 100$ times, and store the weights $(\L_r)_r$ of the second layer for each
run, along with the average matrices and biases %
\begin{equation}
    \bar \W^j =
\frac{1}{R} \sum_{r=1}^R \U^j_r \L_r \qquad \bar \b^j = \frac{1}{R}
\sum_{r=1}^R \b^j_r,\quad\forall\,j \in [N],
\end{equation} 
that combine the
second and third-layer weights and biases for each study $j$ and run $N$, and
average them across runs. We then stack the second-layer weights $(\L_r)_r$
into a tall matrix $\tilde \L \in \RR^{lR \times k}$ on which we perform
sparse non-negative matrix factorization. Namely, we compute $\bar \L \in
\RR^{l \times k}$, the new weight matrix for the second layer, solving
\begin{equation} \bar \L \triangleq \argmin_{\L \in \cC} \min_{\K \in \RR^{lR \times l}}
\frac{1}{2} \Vert \tilde \L - \K \L \Vert_F^2 + \lambda \Vert \K \Vert_F^2,
\end{equation}
where $\cC = \big\{\L \in \RR^{l \times k}, \L \geq 0, {\Vert \l_j
\Vert}_1 \leq 1\:\forall\,j \in [l]\big\}$ and $\lambda$ regulates the sparsity of
$\bar \L$---performance little depends on $\lambda$ provided it leads to finding $\bar \L \D$ with more than 50\% non-zero voxels (see \autoref{sec:cross_val}). Higher $\lambda$ leads to sparser maps with lower performance as brain coverage is reduced, while lower $\lambda$ gives good performances but lower interpretability of the extracted networks. Finally, we compute new weights $\bar \U^j$ for all the
classification heads of the third layer, so that $\bar \W^j \approx \bar \U^j
\bar \L$, from a least-square point of view, for each study $j$. The new model
is then formed of parameters $\D, \bar \L, (\bar \U^j, \bar \b^j)_{j \in [N]}$.
In plain words, we obtain sparse non-negative second-layer weights~$\bar \L$, and define from these weights a new model that is as close as possible to the
ensemble of all learned models $\big\{\D, \L_r, {(\U^j_r, \b^j_r)}_j\big\}_{r
\in [R]}$.

The rows of $\bar \L$ are now interpretable separately, as the non-negative and
sparse constraints have broken the inherent parameter invariance of the
original model. The rows of $\bar \L$ hold the coefficients for combining
resting-state networks held in $\D$ into $l$ multi-study task-optimized networks
$\bar \L \D$ in $\RR^{l \times p}$. We initialize the sparse \acsfont{NMF} algorithm with the weights
$\L_l$ computed in \autoref{sec:nature_rs}, to inject a small prior
regarding final \acs{MSTON} distribution: before running \acsfont{NMF}, those
are set to $ \L_l \D \approx \D_l$, \ie are close to large resting-state
functional networks.

We observed that directly enforcing negativity/sparsity over $\L$ during the
training of the model led to a strong loss in accuracy. Finding a consensus model through a post-hoc ensembling transformation thus
proves to be the right solution for obtaining both performance improvement
\textit{and} interpretability.

\pagebreak
\section{Discussion on the model design}\label{sec:detailed_discussion}

In this section, we discuss various choices made for designing our model and training procedures. To this end, we perform diverse
quantitative and qualitative comparisons of model variants.

\subsection{Understanding the role of task-optimized networks}\label{sec:mston}

We first provide new examples of MSTONs to enlight their properties. Then, we propose several measurements and experiments that allow to better understand how the dimension reduction performed by projecting on multi-study task-optimized networks brings quantitative improvements in decoding.







\begin{figure}[t]
    \centering
    \includegraphics[width=0.33\textwidth]{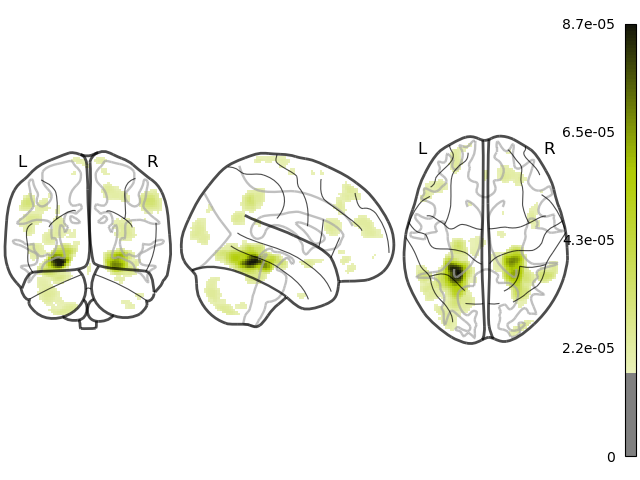}%
    \includegraphics[width=0.33\textwidth]{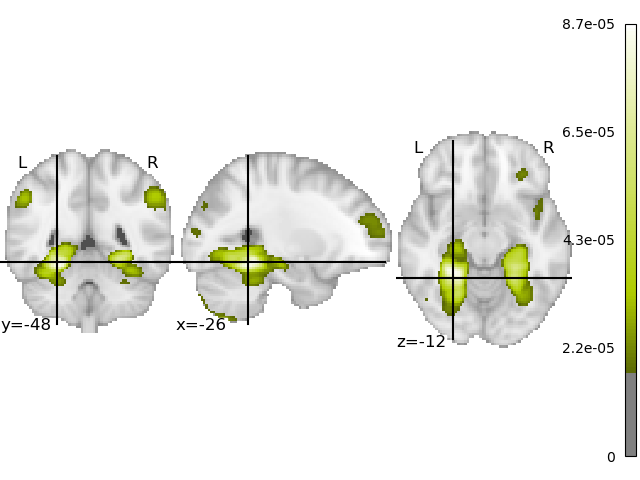}%
    \raisebox{2.7em}{
    \includegraphics[width=0.33\textwidth]{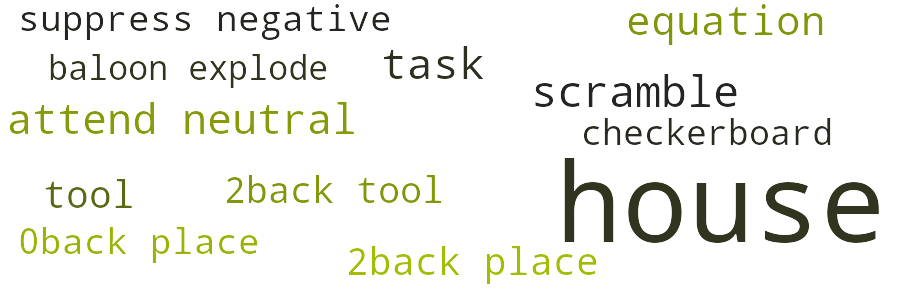}%
    }

    \includegraphics[width=0.33\textwidth]{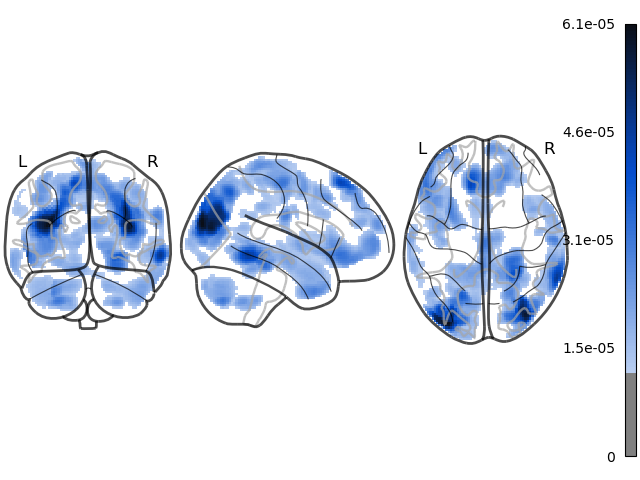}%
    \includegraphics[width=0.33\textwidth]{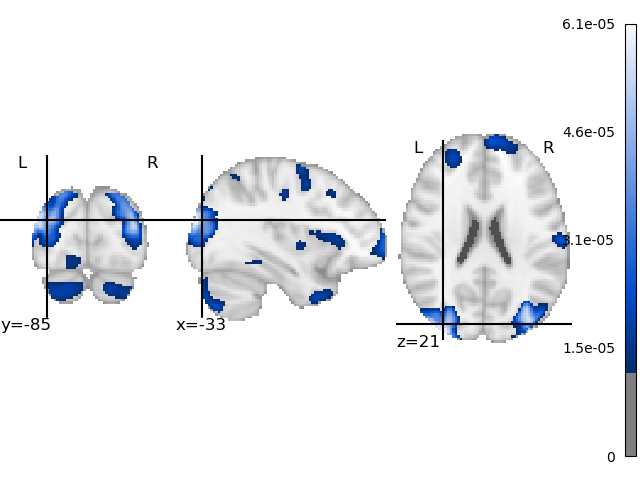}%
    \raisebox{2.7em}{
    \includegraphics[width=0.33\textwidth]{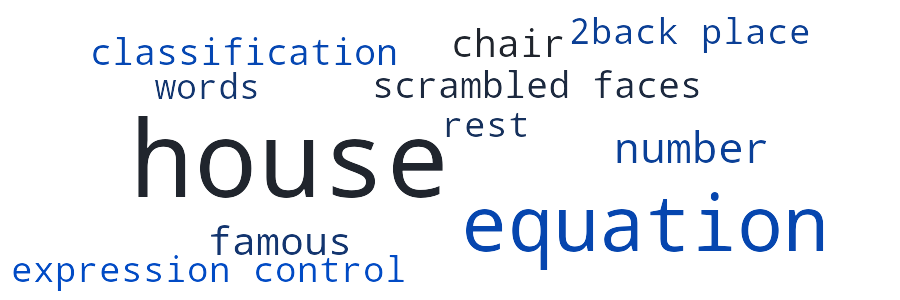}%
    }

    { \sffamily A. MSTONs recruited by the \enquote{house} base condition}

    \includegraphics[width=0.33\textwidth]{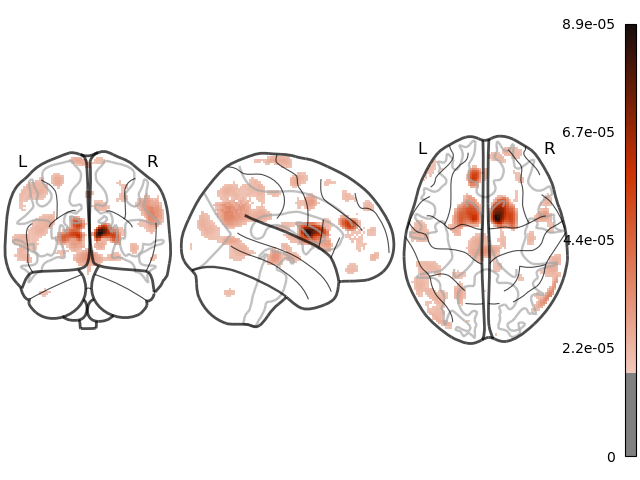}%
    \includegraphics[width=0.33\textwidth]{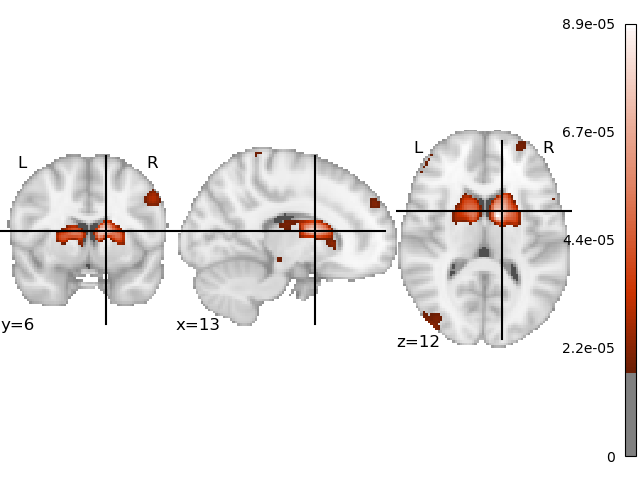}%
    \raisebox{2.7em}{
    \includegraphics[width=0.33\textwidth]{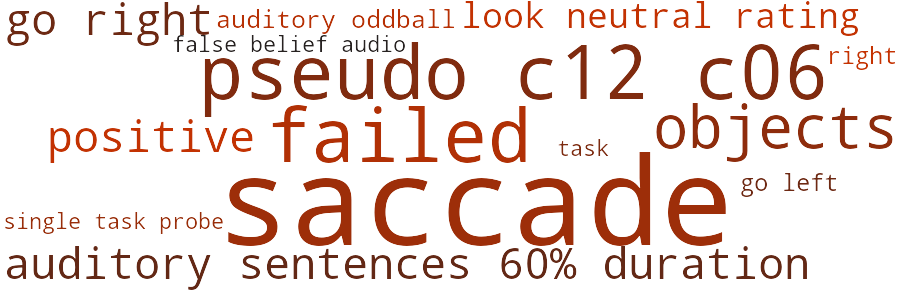}%
    }

    \includegraphics[width=0.33\textwidth]{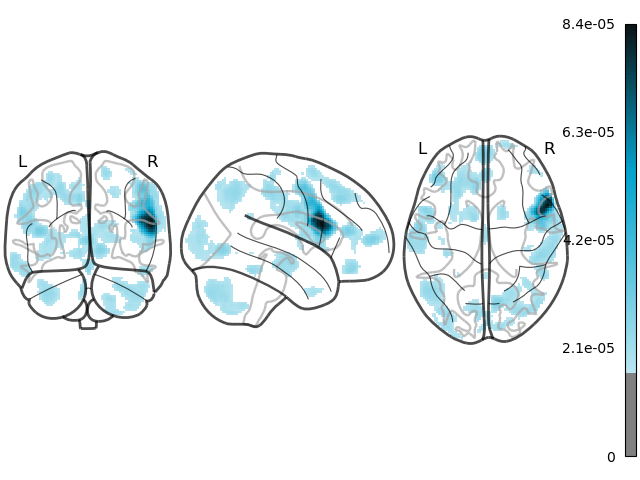}%
    \includegraphics[width=0.33\textwidth]{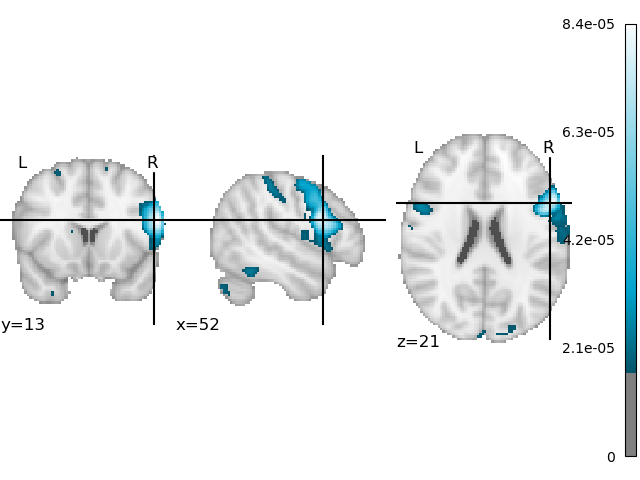}%
    \raisebox{2.7em}{
    \includegraphics[width=0.33\textwidth]{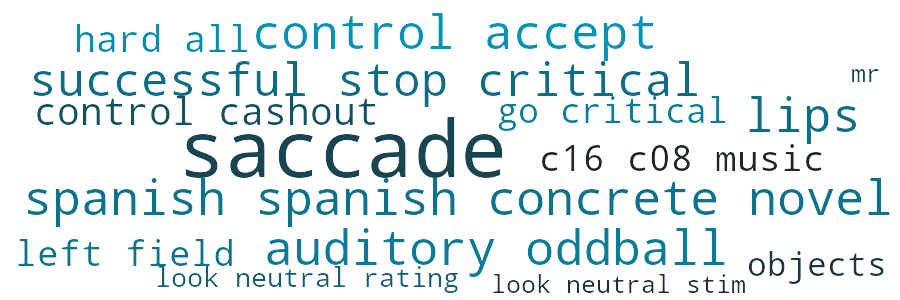}%
    }

    {\sffamily A. MSTONs recruited by the \enquote{saccade} base condition}

    \medskip

    \caption{\textbf{Examples of MSTONs that are activated in many different tasks.}}
    \label{fig:duplicated-components}
\end{figure}

\subsubsection{Other examples of MSTONs}\label{app:mstons_examples}

\autoref{fig:nature_latent} shows a selection of MSTONs that are well associated with relevant clusters of base psychological conditions. Other MSTONs are of interest to discuss the multi-study decoding approach, as we now discuss.


\paragraph{Some base conditions recruit many MSTONs.} We observe that some base
psychological conditions are strongly correlated with many different MSTONs, as exemplified in \autoref{fig:duplicated-components}. The
\enquote{saccade} condition \cite{papadopoulos_brainomics_2017} triggers a
very distributed response of the brain, which is the reason why it appears often
in the word-clouds. The base condition \enquote{house} is in particular part of the HCP Working Memory
\cite{vanessen_human_2012} task. Decoding it versus the other HCP conditions
gives a classification map for which much of the lateral visual cortex is positively
activated, hence the appearance of the word \enquote{house} in the MSTONs that includes a
fraction of these regions.

\subsubsection{Performance of MSTONs on new studies}\label{app:new_studies}

We argue that using the joint objective~\eqref{eq:nature_joint} improves
decoding performance because the data from every study influence the model
weights in both the second layer \textit{and} all components of the third
layer. This can be measured as follows. We compare the performance of learning
task-optimized networks on all studies but a target one, before using the
second layer as a fixed dimension reduction for fitting a decoder from the
target (unobserved) study. Using this technique, information transfer from the
corpus to the new study can only be imputed to the fact that the second layer
has captured a dimension reduction for brain images that is efficient for
decoding in general. In other words, the task optimized networks learned on N -
1 studies form a universal prior of cognition that generalizes to new
paradigms.

\begin{figure}[t]
  \centering
  \includegraphics[width=.8\linewidth]{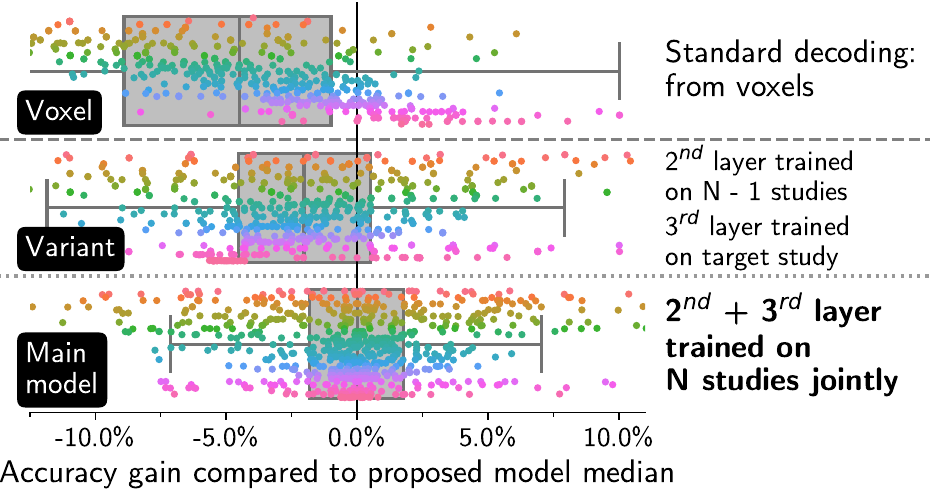}
  \caption{\textbf{Quantitative improvement linked to training the model on the joint
   objective~\eqref{eq:nature_joint}},
    versus improvement linked to transfer in the second-layer only.
     Box plots calculated over 20 random data half-split and all studies.}\label{fig:comparison_method_transfer}
\end{figure}

We observe in \autoref{fig:comparison_method_transfer} that decoding cognitive
processes from externally learned MSTON indeed performs better than decoding
from voxels (3.7\% mean accuracy gain, 67\% experiments with net increase\footnote{Due the fact that half-split folds are overlapping and performance betweens studies are interacting, model comparison experiments are not independent.
This suggests to report the amount of advantageous model comparisons instead of classical null hypothesis testing, that assumes independence of trials.}).
On the other hand, leveraging a low-dimensional representation of
brain images using all studies, including the target one, during
training (1.9\% mean accuracy gain, 75\% experiments with net
increase) performs even better.
This can only be explained by the fact that joint
objective also fosters transfer between the classification heads of the third
layer during training.

\subsubsection{Effect of brain-map dimension reduction}

\begin{figure*}
    \centering
    \includegraphics[width=\textwidth]{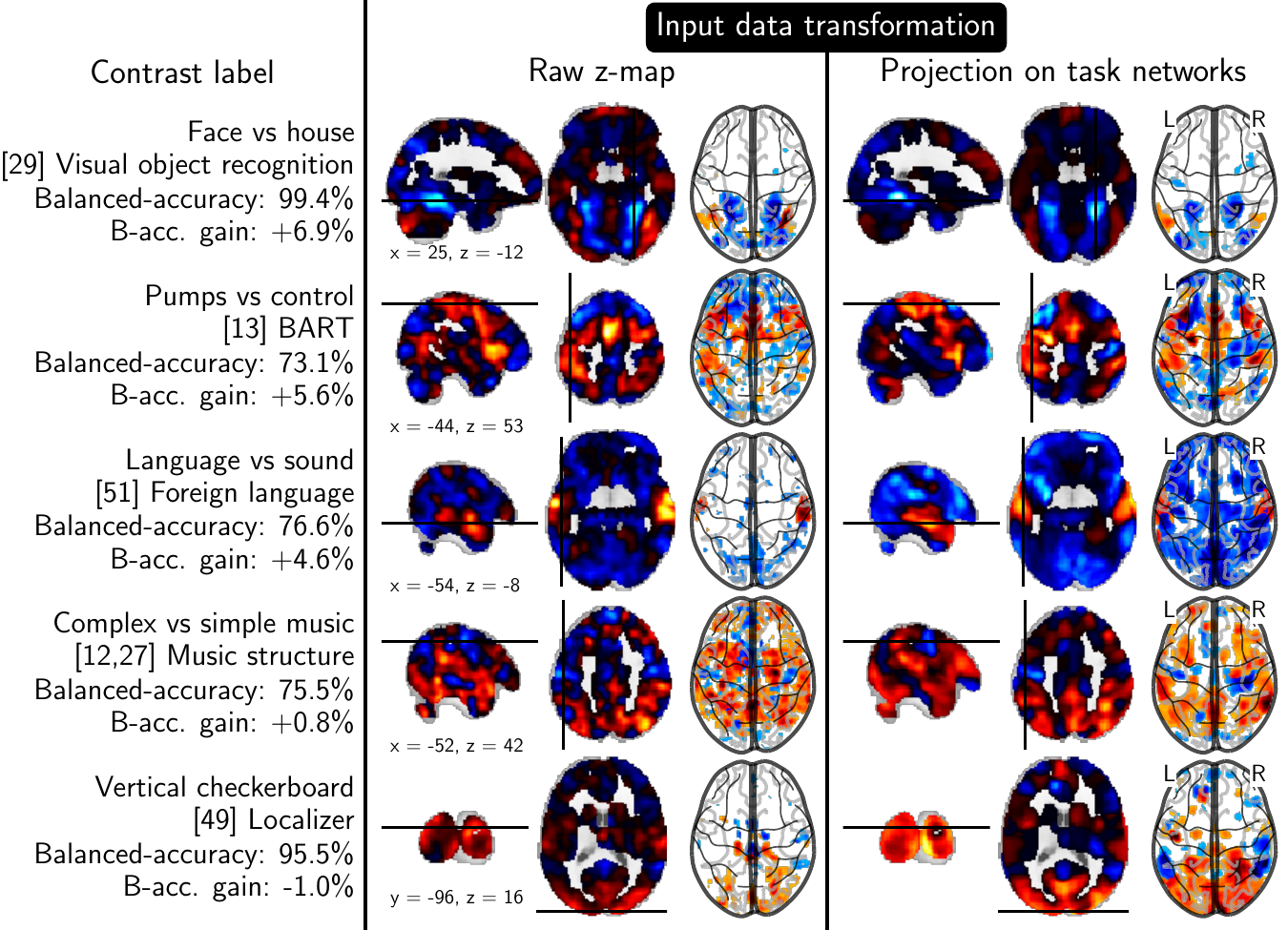}
    \caption{\textbf{Effect of projecting z-maps onto MSTONs.} In a dual perspective to \autoref{fig:nature_classifs}, input data are simplified by the projection onto task-optimized networks, and become easier to classify.}\label{fig:nature_classifs_proj}
\end{figure*}

In a dual perspective, we study the effect of reducing the dimension of the input data with the first two linear layers.
We set $\M = \bar \L \D$ in $\RR^{l \times p}$ to hold the task-optimized networks on each row, and compute, for all input statistical map $\x$ in $\RR^p$, the projection of $\x$ onto $\text{span}(\M)$, namely
\begin{eqnarray}
    \x_{\text{proj}} = \M^\T (\M \M^\T)^{-1} \M \x \in \RR^p.
\end{eqnarray}
$\x_{\text{proj}}$ is thus a denoised, low-dimensional representation of the brain map $\x$, held in the span of the $l$ multi-study task-optimized networks contained in matrix~$\M$. We compare different maps $\x$ to their projection $\x_{\text{proj}}$
in \autoref{fig:nature_classifs_proj}.

\subsection{Fostering transfer learning}\label{sec:transfer_reg}

We now discuss the various way in which we can foster information
sharing across studies in training our multi-layer model.

\subsubsection{The need for objective coupling}

Without modification nor constraint on the second layer output size~$l$, we
cannot expect to observe any transfer learning by solving the joint
objective~\eqref{eq:nature_joint}. Indeed, in the general case where we allow
$l \geq c \triangleq \sum_{j = 1}^N c^j$, we let $(\tilde \V^j, \b^j)_j$ be the unique solutions of the N non-regularized convex
problems~\eqref{eq:classif_loss_red}. We let $\tilde \V \in \RR^{c \times k}$ be the
vertical concatenation of $(\V^j)_j$. We then form the matrices

\begin{align}\label{eq:dummy_sol}
\L &= \begin{bmatrix} \tilde \V \\
 \mathbold{0} \in \RR^{l -c \times k}
\end{bmatrix}
\in \RR^{l \times k}\qquad\text{and} \\
 \begin{bmatrix} \U^1 \\
     \vdots \\
     \U^N
     \end{bmatrix} &\triangleq 
    \begin{bmatrix} \I_c \in \RR^{c \times c}, \mathbold{0} \in \RR^{l - c \times l}
    \end{bmatrix},
\end{align}
where $\I_c$ is the identity matrix of $\RR^{c\times c}$. $\L$ is thus split into row-blocks $(\tilde \V^j)_j$, dedicated to and
learned on \textit{single studies}. It follows from elementary considerations
that the matrices $(\L, (\U^j, \b^j)_j)$ form a global minimizer of~\eqref{eq:nature_joint},
that is formed from the solutions of the \textit{separated}
problems~\eqref{eq:classif_loss_red}. It is therefore possible to find
solutions of~\eqref{eq:nature_joint} for which no transfer occurs.
Two possible modifications of the objective~\eqref{eq:nature_joint} allow to
enforce transfer: Dropout regularization and low-rank constraints,
that we present and compare.

\subsubsection{Dropout as a transfer incentive}

First, as presented in~\autoref{sec:nature_detailed_methods},
we can use dropout between the second layer weight $L$ and the third layer
head weights $\U^j$. Dropout prevents constructions of block-separated solution
of objective~\eqref{eq:nature_joint} similar to the one proposed in~\eqref{eq:dummy_sol}. Indeed, every
reduced sample $\L \D \x_i^j$ fed to the third layer classification head $j$
can see any of his features corrupted by multiplicative noise $\M_L$ during
training. This pushes the model to capture information relevant for all studies
in every activation of the second layer. In other word,
the projection performed on any task-optimized network $\l_h \D$, for $h \in
[l]$ should be relevant for decoding every study. This fosters transfer
learning as~$\L$ carries multi-study aggregated information at the end of training,
unlike in~\eqref{eq:dummy_sol}.

\subsubsection{Transfer through low-rank constraints/penalty}\label{sec:penalty_transfer}

A second approach to transfer is to force the matrices
\begin{equation}\label{eq:def_u_v}
    \V \triangleq 
    \begin{bmatrix} \V^1 \\
        \vdots \\
        \V^N
        \end{bmatrix}
    \triangleq
    \begin{bmatrix} \U^1 \\
        \vdots \\
        \U^N
        \end{bmatrix} \L,
\end{equation}
formed of the parameters of the joint objective~\eqref{eq:nature_joint} to be
\textit{low-rank}. In this case, the subspace of $\RR^{c \times k}$ in which
$\V$ evolves is strictly smaller than $\RR^{c \times k}$, and we cannot always
find a global minimum of the joint objective~\eqref{eq:nature_joint} formed
with the solutions $\tilde \V$ of the separate
objectives~\eqref{eq:classif_loss_red}, as we did in the
construction~\eqref{eq:dummy_sol}. As a consequence, the data from studies
truly influence the solutions $(\L, {(\U^j, \b_j)}_j)$
of~\eqref{eq:nature_joint}, and transfer is theoretically possible.

The low-rank property may be enforced in two ways.
First, we may set it as a hard constraint, setting $l < c$ in the joint
objective~\eqref{eq:nature_joint}. This is in practice what we do when
selecting $l = 128$, as $c = 545$ in our experiments.

Alternatively, following~\cite{srebro_maximum-margin_2004}, we may resort to a
convex objective function parameterized by $\V$ in $\RR^{c \times k}$, that
penalizes the rank of $\V$. We learn $\V^j$ in $\RR^{c^j \times k}$ for all study $j$
in $[N]$ solving the joint objective
\begin{align}
    \min_{(\V^j, \b^j)_j} &- \sum_{j=1}^N
    \frac{(n^j)^\beta}{n^j} \sum_{i=1}^{n^j}    
    \Big(l^j_{i, y_i}(\V^j, \b^j) \\ 
     &- \log(\sum_{k=1}^{c^j} 
    \exp l^j_{i, k}(\V^j, \b^j)) \Big) \notag \\ &+ \lambda \left\Vert \begin{bmatrix} {\V^1}^\top \dots {\V^N}^\top \end{bmatrix} \right\Vert_\star,
    \label{eq:nature_joint_trace}
\end{align}%
where $\Vert \V \Vert_\star$ is the nuclear norm of $\V$, defined as \linebreak
$\sum_{i=1}^{\min(c, k)} \sigma_i(\V)$, where $(\sigma_i(\V))_i$ are the singular values
of $\V$. The nuclear norm is a convex proxy for the rank of matrix~$\V$. As a
consequence, the rank of the solution decreases from $\min(c, k)$ to $0$ as
$\lambda$ increases. The objective~\eqref{eq:nature_joint_trace} is solvable
using proximal methods, \eg \acs{FISTA} \cite{beck_fast_2009}. 
However,  these methods become unpractical when~$c$ becomes large---it
requires to perform a $c \times c$ singular value decomposition at each iteration. 
Fortunately, there exists a non-convex objective~\cite{rennie_fast_2005},
amenable to stochastic gradient descent~\cite{bell_lessons_2007}, that
includes the solution of~\eqref{eq:nature_joint_trace} as a minimizer. It is
obtained by setting $l = \max(x ,k)$ and adding $\ell_2^2$ penalties to the
objective~\eqref{eq:nature_joint}:
\begin{align}\label{eq:nature_joint_l2}
    \min_{\substack{\L \in \RR^{l \times k}\\
    (\U^j, \b^j)_{j}}} \!\!\! &- \sum_{j=1}^N
    \frac{(n^j)^\beta}{n^j} \sum_{i=1}^{n^j}    
    \Big(l^j_{i, y_i}(\U^j, \b^j, \L) \\
    &- \log(\sum_{k=1}^{c^j} \exp l^j_{i, k}(\U^j, \b^j, \L)) \Big) \\& +
    \frac{\lambda}{2} \Big( \Vert \L \Vert_F^2 
    + \sum_{j=1}^N \Vert \U^j \Vert_F^2 \Big).
\end{align}%
We solve this objective using \textit{Adam}, similarly to the main method. It
is possible to continue using dropout in between the first and second layer
while enforcing~$\V$ to be low-rank---this can then be understood as a regularization technique through feature noising~\cite{wager_dropout_2013}.

\subsubsection{Empirical comparison of transfer penalties}\label{sec:transfer_penalties}

\begin{figure}
    \centering
    \includegraphics[width=.8\linewidth]{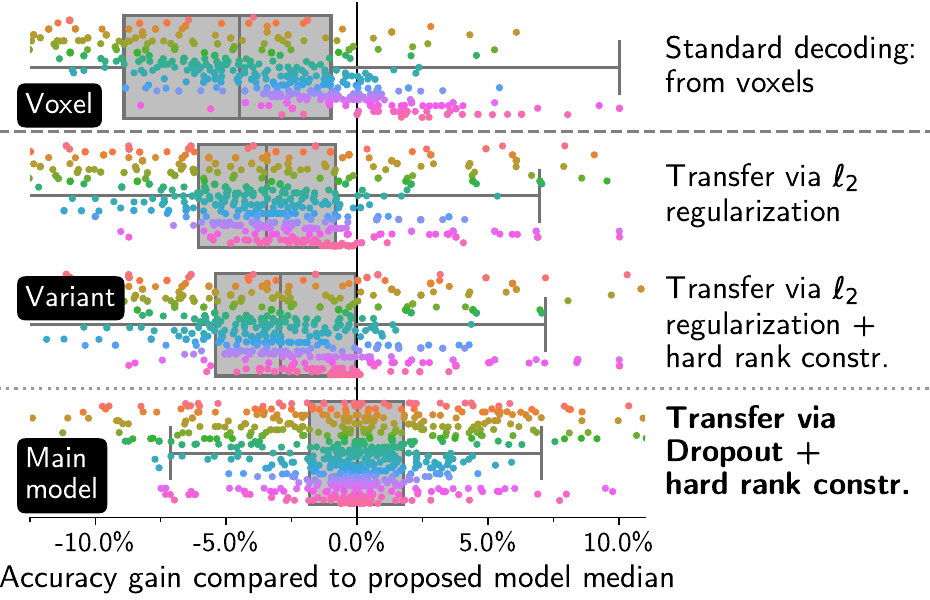}
    \caption{\textbf{Quantitative comparison of transfer inducing regularizations.} Dropout with hard-rank constraints outperforms $\ell_2$ regularization with and without hard-rank constraints. Box plots calculated over 20 random data half-split and all studies.}\label{fig:l2}
\end{figure}

\paragraph{Dropout versus $\ell_2$.} Both the dropout and low-rank approaches are a priori competitive to foster
transfer learning. Our final method uses a combination of both, as it enforces
a hard low-rank constraint and uses dropout. This choice was motivated by a first experiment, summarized in \autoref{fig:l2}.
We compare three regularization variants by measuring the improvement due to hard
low-rank constraints and the difference between dropout and $\ell_2$. The three
estimators use input dropout ($p = 0.25$).
The first two estimators use $\ell_2$
regularization. Dropout between layer 2 and 3 is initialized to $p = 0.75$ in
the third estimators.  The first estimator does not use a hard-rank constraint
($l = c = 545$), while others use $l = 128$.\footnote{The reported $\ell_2$
accuracy gain is larger than its actual performance when $\lambda$ is set with
cross-validation, as we take the highest performing $\lambda$ on the
\textit{test} sets. Symmetrically, we may slightly improve results by setting
dropout rates using cross-validation---we choose not to, to avoid the fragility
of cross-validation in neuro-imaging~\cite{varoquaux_cross-validation_2018}.} We
observe that forcing $\V$ to be low-rank is beneficial (0.7\% mean accuracy
gain, 72\% experiments with net increase) in the absence of dropout, and that
dropout regularization performs significantly better than low-rank inducing
$\ell_2$ penalties (2.7\% mean accuracy gain, 79\% experiments with net
increase). This justifies using dropout regularization.

\begin{figure}
    \centering
    \includegraphics[width=\linewidth]{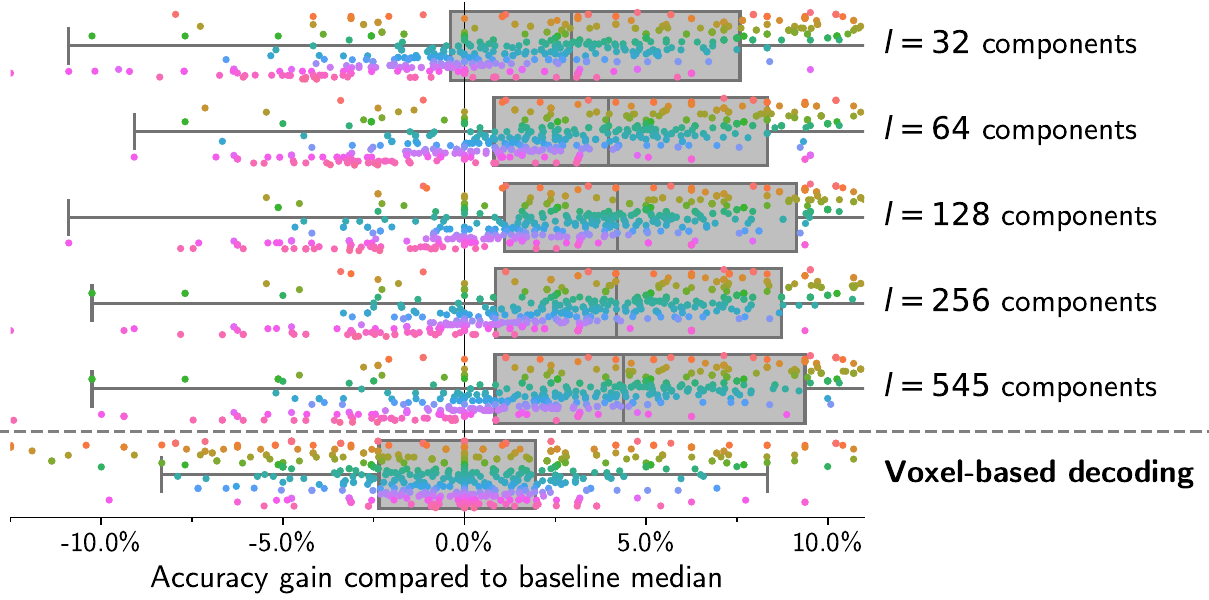}
    \caption{\textbf{Performance of multi-study decoding for varying second layer width $l$.}}\label{fig:varying_l}
\end{figure}

\paragraph{Low-rank constraints and second-layer width.} With dropout, the
performance of multi-study decoding varies with the size of the latent space $l$,
as displayed in \autoref{fig:varying_l}. The performance reaches a plateau at $l
\approx 128$. Setting a high $l$ results in more scattered networks, so that
different but similar MSTONs may be recruited to decode the same psychological
condition (see examples in \autoref{fig:duplicated-components}). Choosing a low $l$ leads
to slightly worse performances but more interpretable components. We therefore
use $l = 128$, as if offers the best performance/interpretability trade-off.

\paragraph{First-layer width.} Some previous work \cite{dadi2019fine} studies the impact of
projecting brain signal onto $k$ functional units, for varying $k$ and different
fMRI analysis tasks. The conclusion of this work applies here: setting a high
$k$ ensures the best performances. We use $k=465$ grey-matter components
extracted from $512$ full-brain components due to constraints in
training---higher $k$ may be used in future work.

\begin{figure}[h]
    \centering
    \includegraphics[width=\linewidth]{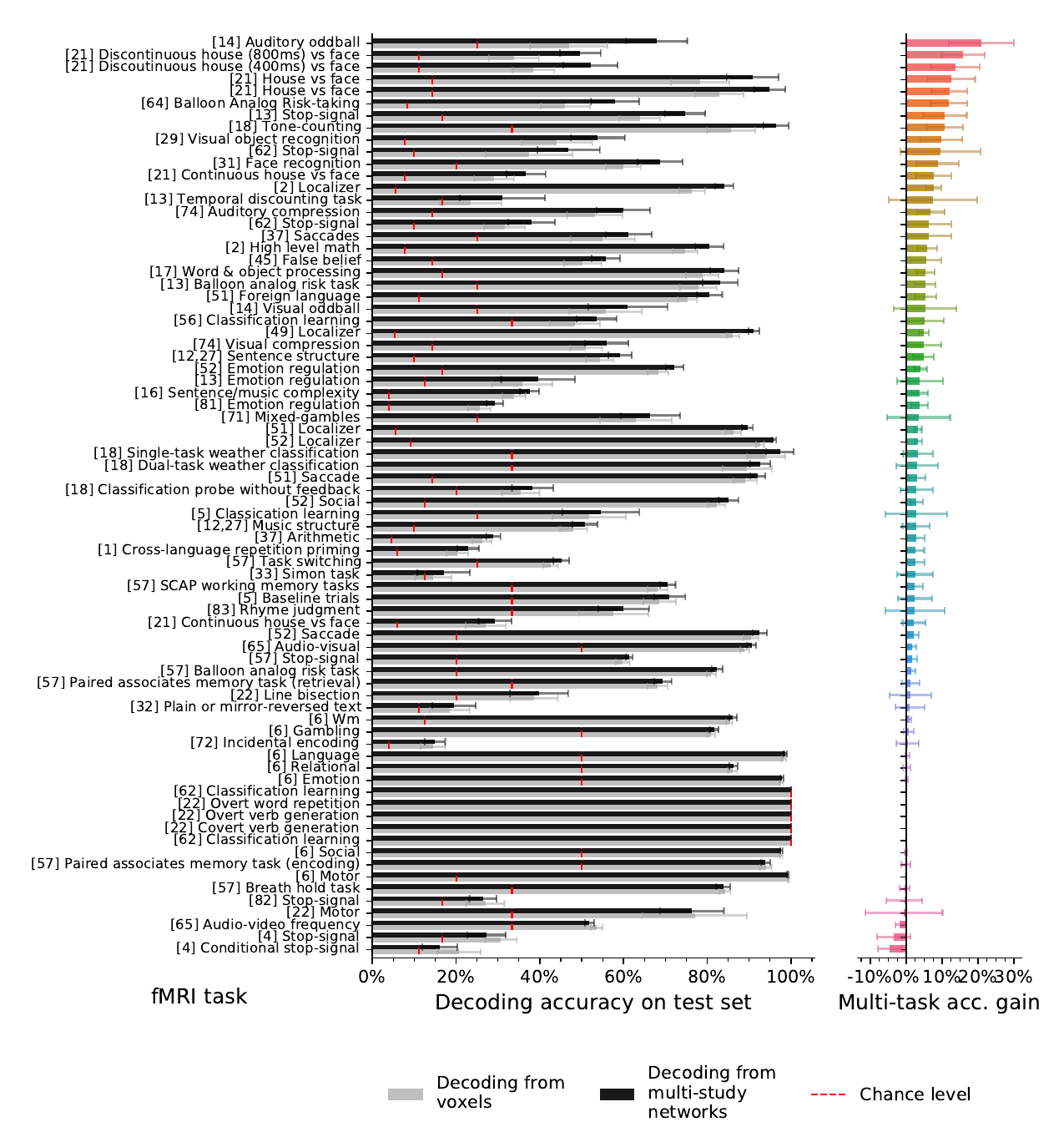}

    \caption{\textbf{Performance of multi-study multi-task decoding,} versus single-study single-task decoding from resting-state functional units. Numbers are reported in \autoref{table:accuracy_task}.}
    \label{fig:per-task-quantitative}

\end{figure}

\subsection{Multi-study multi-task decoding}\label{app:per-task}

We have validated the multi-study decoding approach in a \textit{per-site} setting, in which each study defines a single decoding task. Some studies include different fMRI tasks: we can also use each of these tasks to define a single decoding problem, and perform \textit{multi-study multi-task decoding}. To evaluate this approach, we use the task annotations from the 35 studies of our corpus and obtain 76 classification tasks to be solved simultaneously. We compare the performance of the three-layer model, versus single-task decoding from the resting-state functional units. We use the exact same architecture as for multi-study training.

Results are displayed in \autoref{fig:per-task-quantitative}. Multi-task
training brings an improvement for 62/76 tasks. Quantitatively, the mean
improvement is lower than the one obtained for within-study decoding ($+3.9\%$
vs $+5.8\%$). This was expected, as the average chance-level in within-task
decoding is higher than in within-study decoding. Using multi-task or multi-site modelling should depend on the purpose of the study.

\subsection{Interpretability incentives}\label{sec:initialization}

\begin{figure}
    \centering
    \includegraphics[width=.96\linewidth]{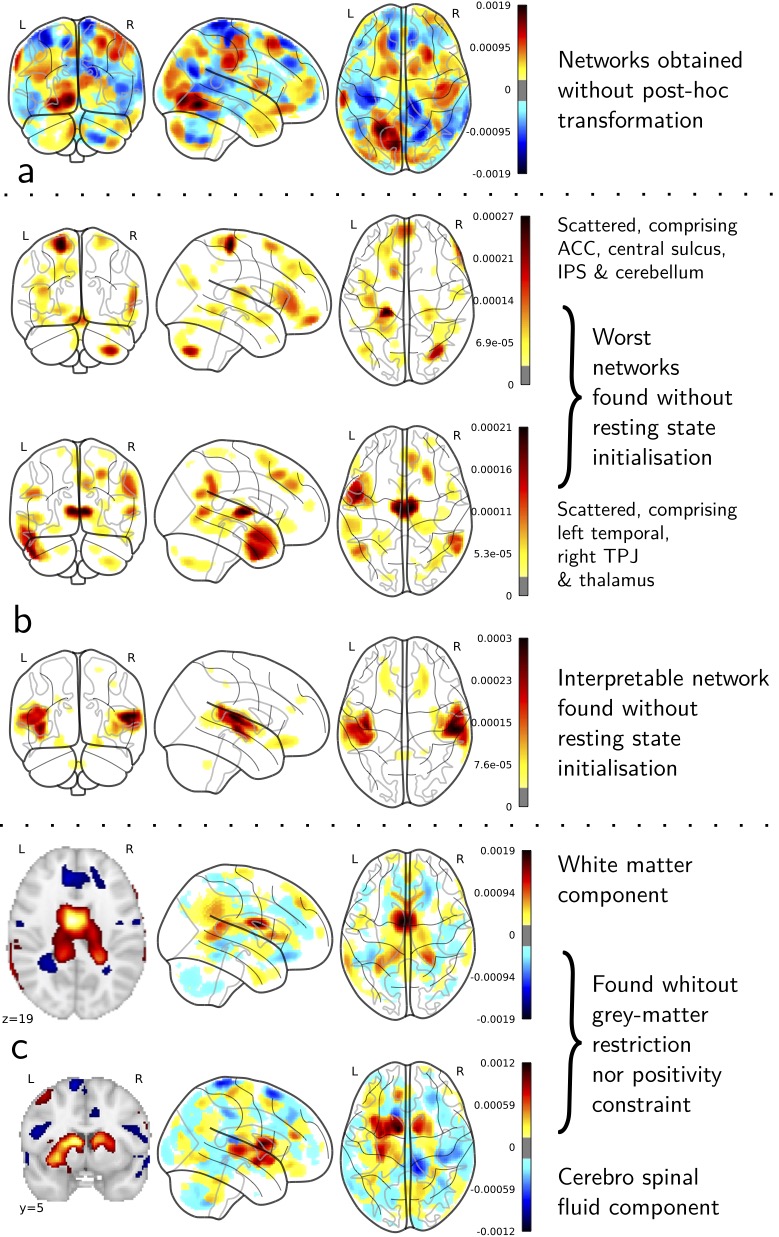}
    \caption{\textbf{Effects of components selection}. Without post-hoc transformation \textbf{(A)} resting-state based initialization \textbf{(B)}
    and grey matter components selection \textbf{(C)}, some task-optimized networks may be hard to interpret or not relevant from a cognitive perspective.}\label{fig:bad_components}
\end{figure}

A core feature of our approach is model interpretability. Three
aspects allow to find cognitive meaningful task-optimized networks.
First, the initial first layer, learned on resting-state data, coarsens the
resolution of networks in a way adapted to typical brain signals. Second, we
compute a consensus model, so
that the task-optimized network loadings held in $\L$ are non-negative and
interpretable. Third, we initialize the second-layer weights so that
$\L_{\text{init}} \D$ corresponds to resting-state functional networks $\D_l$, coarser than $\D$. This
initialization is used both during the training phase and the consensus phase.

\begin{figure}
    \centering
    \includegraphics[width=.8\linewidth]{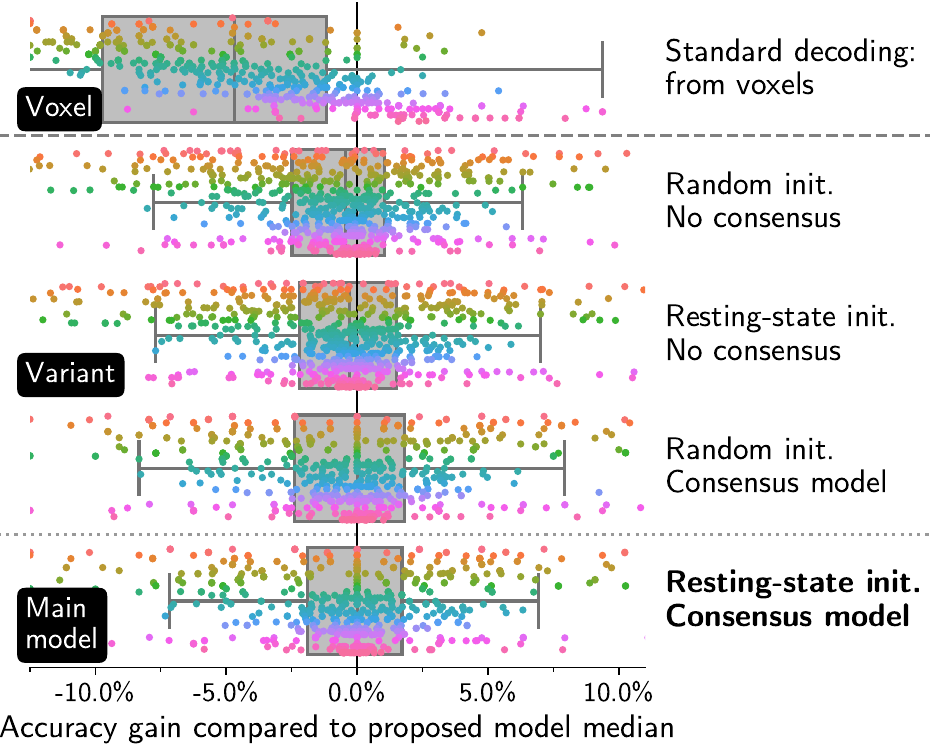}
    \caption{\textbf{Quantitative improvement linked to ensembling and
    resting-state initialization.} Box plots calculated over 20 random data
    half-split and all studies.}\label{fig:posthoc}
\end{figure}

\paragraph{Consensus model and resting-state initialization.} In \autoref{fig:posthoc}, we measure the  
quantitative effects of the two later factors on decoder accuracy. Learning a consensus model using sparse \acs{NMF} is crucial for finding interpretable direction in the span of
$\L$.  Without this refinement, the directions we obtain are similar to the one displayed in \autoref{fig:bad_components}A, and are less interpretable. Both the consensus phase and the resting-state initialization
contributes positively to the model decoding performance (0.6\% mean accuracy gain, 66\% experiments with net increase). We attribute this improvement to an ensembling effect similar to the benefits of bagging~\cite{breiman_bagging_1996}, as the
final model summarizes 100 training runs on the same data, with different
random seeds, and to the fact that resting-state networks form a good prior for task-optimized network.

Qualitatively, we show examples of three components found
without resting-state initialization in \autoref{fig:bad_components}B. Two of those are
scattered networks, that capture various connected components whose co-occurrence is
not interpretable: those components are likely artifacts due to random initialization.
Using resting-state initialization finds such networks much less frequently. It
remains interesting to note that most of the components found without
resting-state based prior bear cognitive meaning, similar to the third components displayed in \autoref{fig:bad_components}B.

\subsubsection{Effect of selecting grey-matter components}\label{sec:grey_matter}

\begin{figure}
    \centering
    \includegraphics[width=.8\linewidth]{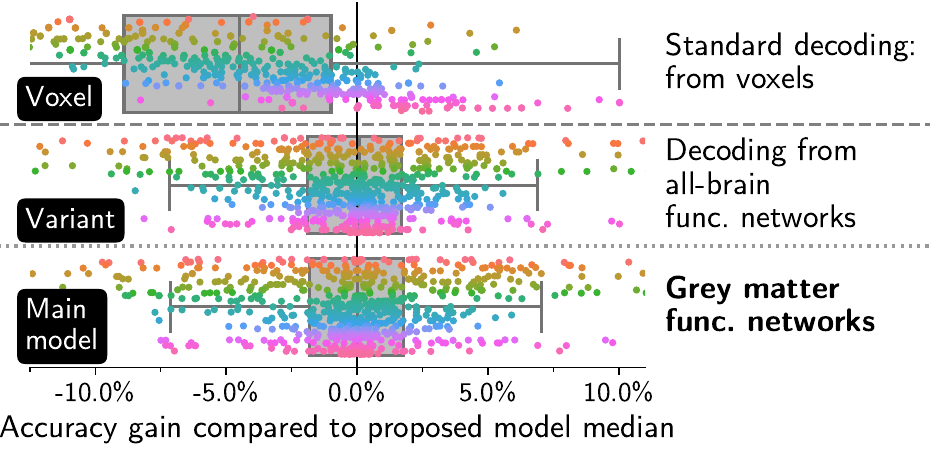}
    \caption{\textbf{Quantitative improvement linked to working with a
    grey-matter mask.} Working with functional networks located in the grey
    matter only do not have a significant impact on performance. Box plots
    calculated over 20 random data half-split and all
    studies.}\label{fig:comparison_method_gm}
\end{figure}

We project data onto a subset of 465 out of 512 functional networks learned on
\acs{HCP} resting-state data, selecting the networks that intersect with
an anatomical grey-matter mask. This avoids finding \acs{MSTON}s that are distributed or formed with non grey-matter regions. In \autoref{fig:bad_components}C, we show that
without those precautions, our model finds networks located in the white matter
and the cerebro-spinal fluid zones. Quantitatively (\autoref{fig:comparison_method_gm}), as expected, performing
classification from grey-matter components only brings a non-significant
performance loss (0.03\% median accuracy gain).

\subsection{Effect of variational dropout and batch normalization}\label{sec:var_dropout}

\begin{figure}
    \centering
    \includegraphics[width=.8\linewidth]{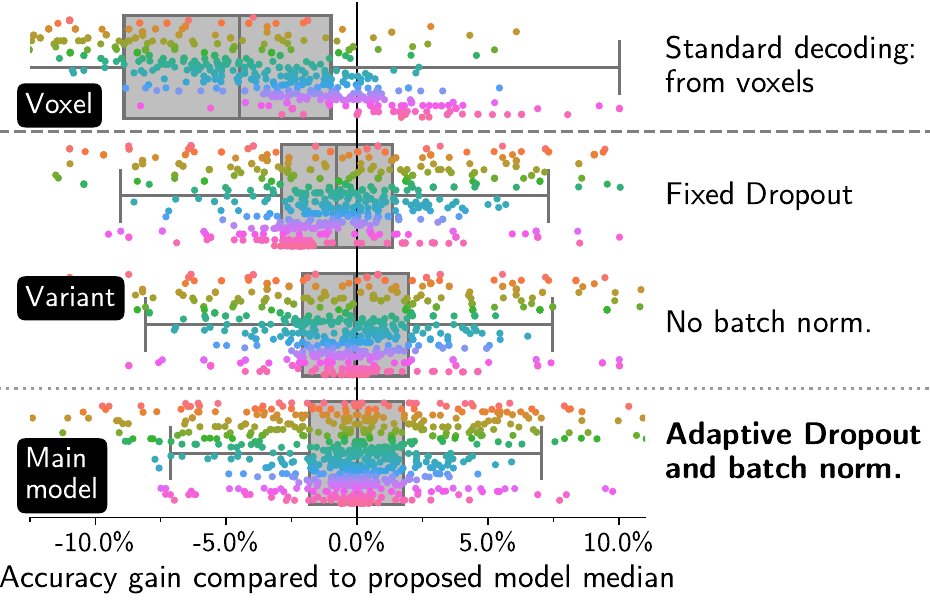}
    \caption{\textbf{Batch normalization and adaptive variational dropout both have a beneficial impact on classification accuracy of the final learned decoder.} Box plots calculated over 20 random data half-split and all studies.}\label{fig:comparison_method_dropout}
\end{figure}

We introduced variational dropout and batch normalization in the training
procedure of our algorithm. \autoref{fig:comparison_method_dropout} shows
that it is indeed beneficial. Variational dropout brings a mean accuracy gain of 0.7\% (64\% experiments with net increase) compared
to binary dropout; batch normalization benefit is smaller but positive (0.1\% mean accuracy gain, 55\% experiments with net increase), and allows
faster training---in line with its original purpose~\cite{ioffe_batch_2015}.

\subsection{Stronger improvement for smaller studies}
\label{app:small}

To verify the finding of \autoref{fig:training_curves} and evaluate the impact of training-size on multi-study decoding, we perform the following experiment. We restrict the study corpus to studies with more
than $30$ subjects, train the three-layer model on $15$ subjects from each
study, and evaluate its performance on the remaining population. We repeat this
experiment $20$ times. 

\begin{figure}[h]
    \centering
    \includegraphics[width=\textwidth]{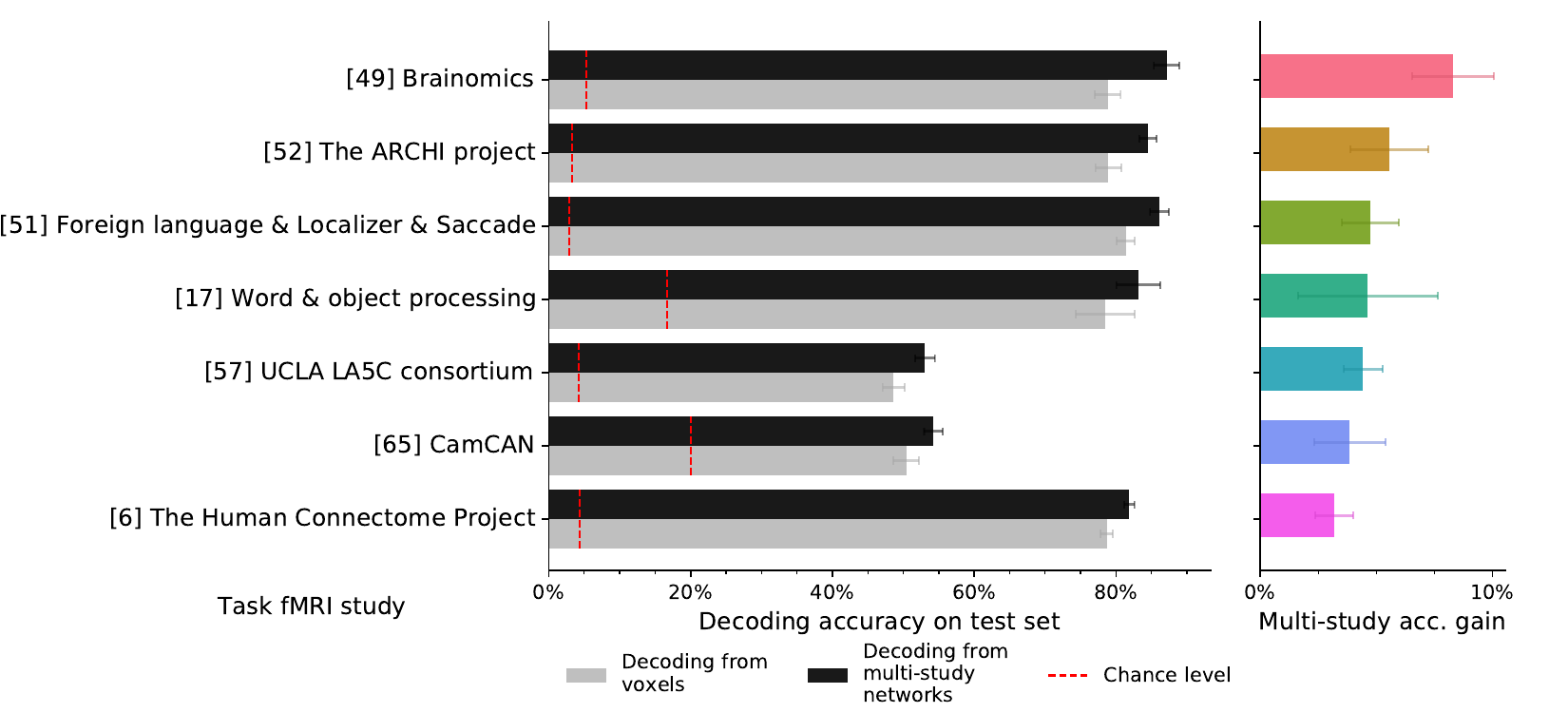}
    \caption{\textbf{Performance of multi-study decoding with 15 training subjects per study.}}
    \label{fig:quantitative_subjects}
\end{figure}

We report results in \autoref{fig:quantitative_subjects}. Transfer learning is
positive for all studies (mean accuracy gain $+4.8\%$.). This includes
studies with a large complete cohort, for which transfer learning is uneffective when considering all
available subjects (e.g. HCP, \autoref{fig:nature_quantitative} and data from
the UCLA consortium). The multi-study approach is therefore particularly efficient
for studies with less than $30$ subjects, that are still the most common in the literature.

\subsection{Effect of decoding difficulty}\label{app:decoding_difficulty}

\begin{figure}[t]
    \centering
    \includegraphics[width=\linewidth]{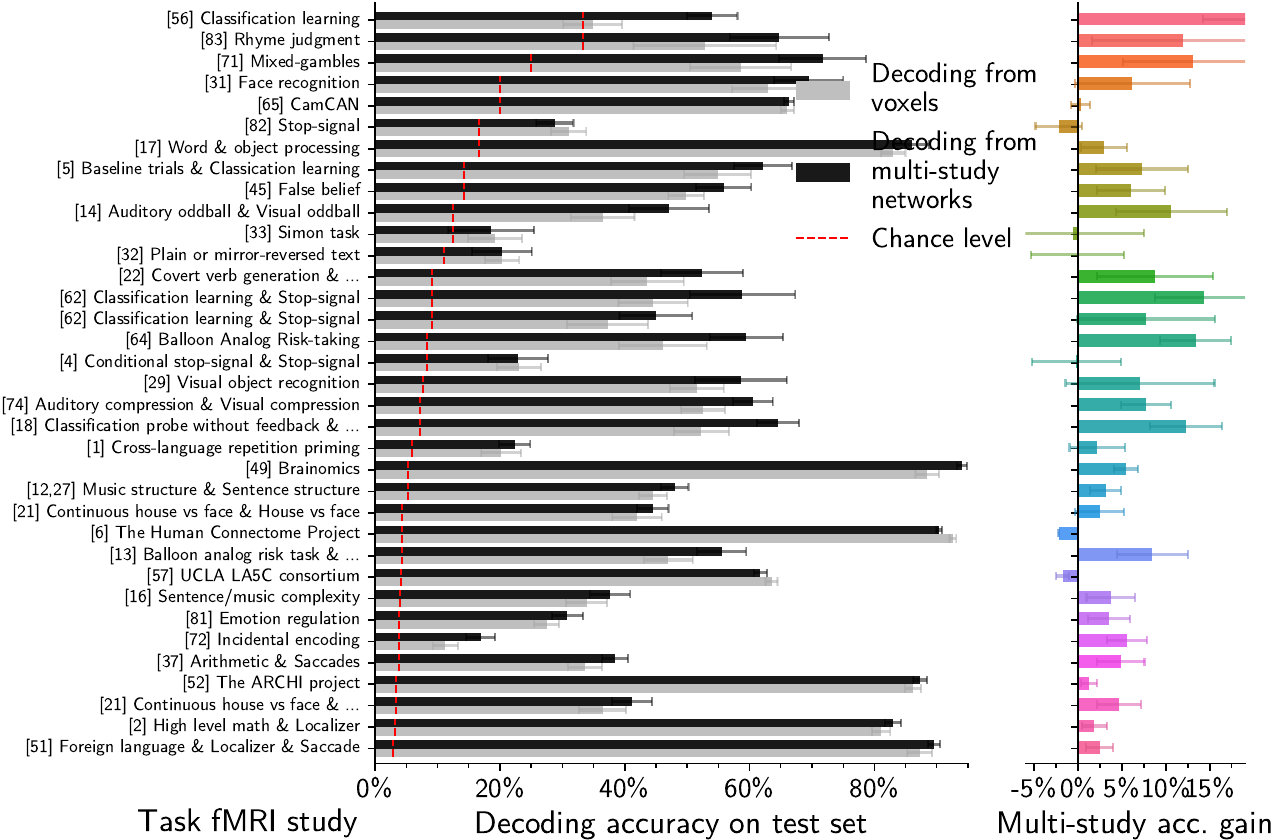}
    \caption{\textbf{Performance improvement of multi-study decoding} vs voxel-level decoding, \textbf{sorted by the chance level} of the decoding task of each study.}
    \label{fig:sorted_quantitative}
\end{figure}

We investigate how the difficulty of a given decoding task (provided by a single study) influences the performance improvement due to multi-study decoding. For this, we report in \autoref{fig:sorted_quantitative} the same numbers as in \autoref{fig:nature_quantitative}, sorting studies by their chance level: lower chance level means \enquote{harder} decoding tasks, as contrasts must be selected in larger sets. We observe a slight tendency of higher improvement for easier tasks, although no strong conclusion may be drawn.

\subsection{Effect of study weights}\label{sec:study_weight}

\begin{figure}
    \centering
    \includegraphics[width=.4\linewidth]{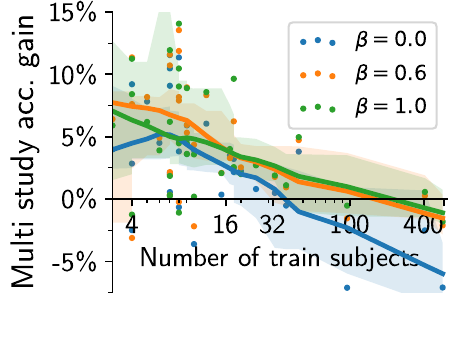}%
    \includegraphics[width=.4\linewidth]{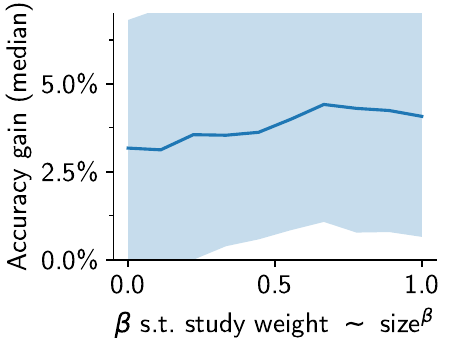}
    \caption{\textbf{Impact of changing the study weight in the joint objective}. Giving more weight ($\beta \to 1$) to large studies prevents negative transfer learning but may reduce overall performance. Small studies should not be given too much weight ($\beta \to 0)$, as this voids the benefits of jointly training over bigger studies. An intermediary $\beta = 0.6$ gives the best performances. Error bars calculated over 20 random data half-split and all studies.}\label{fig:study_weight}
\end{figure}

Our model learns the second and third layer weights by solving 
\begin{align}
    \min_{\substack{\L \in \RR^{l \times k}\\
    (\U^j, \b^j)_{j}}} \!\!\! &- \sum_{j=1}^N
    \frac{(n^j)^\beta}{n^j} \sum_{i=1}^{n^j}    
    \Big(l^j_{i, y_i}(\U^j, \b^j, \L) \\
    &- \log(\sum_{k=1}^{c^j} \exp l^j_{i, k}(\U^j, \b^j, \L)) \Big),
\end{align}%
in which the many studies can be given various weights. At one extreme, we may consider
that all studies of the corpus should be weighted the same, which amounts to setting
$\beta = 0$ in~\eqref{eq:nature_joint}. At the opposite, we can consider that
each brain map from each study should have the same importance, which amounts to
setting $\beta = 1$. As \autoref{fig:study_weight}B shows, it is beneficial to set an
intermediary $\beta$, typically $\beta = 0.6$. On the one hand, we want to
give the smallest study of our corpus a non negligible importance; on the other
hand, we want the large studies to
remain more weighted than the smaller ones, as they should provide more accurate information. Our reweighting amounts to
giving every study $j$ an ``effective sample size''
\begin{equation}
    n^j_{\text{eff}} = \sum_{i=1}^N n^i \frac{{n^j}^\beta}{\sum_{i=1}^N {n^i}^\beta},
\end{equation}
that is larger than the true sample size for smaller studies and smaller for larger studies.
We observe on \autoref{fig:study_weight}A that the negative transfer
learning endured by large-study decoders such as \acs{HCP} and \acsfont{LA5C}
reduces as these studies are given more weight ($\beta \to 1$). On the other
hand, the performance on small datasets slightly reduces for $\beta > 0.6$. It
also reduces for low $\beta$, hinting at the importance of using large studies for
improving small studies decoding.

We thus have provided justifications for all the technical design choices made in
training our decoding model: regularization, joint training, training refinements,
choice of study weights.

\subsection{Comparison with earlier work}\label{app:comparison}

We proposed a proof-of-concept, smaller-scale and harder to interpret multi-study decoding approach in \cite{mensch2017learning}. This earlier work already relies on a three-layer linear model, with joint training of the second and third layer. Beyond its extended cognitive neuroscience point-of-view, the present work strongly improves the multi-study decoding methods and results.

\paragraph{Model interpretability.} From a methodological point of view, \cite{mensch2017learning} fail short of providing a principled way for interpreting results and extracting meaningful task-optimized networks, as those outlined in \autoref{fig:nature_latent}. Their approach yields networks akin to \autoref{fig:bad_components}A, which are not relevant from a cognitive perspective. A template-extracting approach that clusters the low-dimensional brain map representations is proposed; yet it remains exogenous to the model and does not perform convincingly. The consensus post-hoc transformation method we propose in this work addresses the issue of interpretability and finds cognitive directions that efficiently capture mental state information. As \autoref{fig:comparison_method_transfer} shows, these meaningful networks can be used as a cognitive atlas for improving decoding on newly acquired datasets, without joint training. Consensus through matrix factorization of the model weights also increases model performance (\autoref{fig:comparison_method_transfer}).

\paragraph{Architecture, constraints, training.}The functional atlases used as a first-layer by \cite{mensch2017learning} are smaller (up to 256 components) and not constrained to be non-negative. As we discovered, enforcing non-negativity of the first layer $\D$ and the second layer $\L$ (after ensembling) is crucial to interpret the prediction of the model. Using a larger functional atlas extracted from resting-state data ensures that no information is lost when reducing the dimension of brain maps. Initialization of the second-layer with resting-state information increases the model performance (\autoref{fig:posthoc}), as well as the use of variational dropout \cite{kingma_variational_2015} and batch normalization \cite{ioffe_batch_2015} (\autoref{fig:comparison_method_dropout}).

\paragraph{Data and validation.}
\cite{mensch2017learning} pool only the results of 5 studies, which prevents the observation heavy transfer effects, and the extraction of broadly-valid cognitive directions. The present work validates the approach on 7 times more studies, proving that our multi-study approach is valid beyond proof-of-concept, and truly promising for the neuroscience community. To better explain the transfer of information across studies, we compare several transfer approaches (convex models, low-rank constraints, stochastic regularization: see \autoref{sec:transfer_reg}), and assess how classification maps are affected by the use of task-optimized network (Figs \ref{fig:nature_classifs}, \ref{fig:nature_dendro} and \ref{fig:nature_classifs_proj}); this endeavor is missing in earlier work.

\pagebreak

\section{Reproduction details and tables}\label{sec:data_corpus}

In this last section, we detail our experiment pipeline, the numerical parameters needed for reproducing this study, and the sources from which we obtained our corpus of studies.

\subsection{Software and parameters}\label{sec:cross_val}

We used \textit{nilearn}~\cite{abraham_machine_2014} and
\textit{scikit-learn}~\cite{pedregosa_scikit-learn:_2011} in our experiment
pipelines, the stochastic solver from~\cite{mensch_stochastic_2017} to learn
resting state dictionaries and \textit{pytorch}~\cite{paszke2017pytorch} for
model design and training. The \textit{cogspaces} package that we have published
provides the multi-scale resting-state dictionaries extracted from \acs{HCP}, as
those are costly to learn. It also provides the reduced representations of the
data from the 35 studies we consider.

\paragraph{General cross-validation scheme.}For every validation experiment and
comparison, we perform 20 half-split of all data. Name\-ly, we consider half of
the subjects of every study for training, and test the decoder on the other
half. As two studies~\cite{ds017} share subjects, we also ensure that no single
subject appears in both the training and the test sets across studies.

\paragraph{Baseline parameter selection.} We cross validate the $\lambda$
parameter for the baseline multinomial regression classifiers, on a grid
\begin{equation}
    \{10^i, i = \{-3, -2, -1, 0, 1, 2, 3\} \}.
\end{equation}
\paragraph{Dropout rate.}We use a dropout rate of $p = 0.25$ in between the
first and second layer and initialize study-specific dropout rates with $p =
0.75$ in between the second-layer and third-layer classification heads (\ie we
set $\alpha = \frac{p}{1 - p}$ in variational dropout). \paragraph{Resting-state
dictionaries.}We obtain the 512-components and 128 components resting-state
dictionaries by choosing $\lambda$ on a grid
\begin{equation}
    \{10^i, i = \{-5, -4, -3, -2, -1, 0, 1\} \},
\end{equation}
so to obtain components that cover the whole brain with minimal overlap.
\paragraph{Consensus phase.}We run the training procedure 100 times with different random seeds. We set $\lambda = 10^{-4}$, so as to obtain $80 \%$ sparsity. We tried $\lambda \in \{10^{-5}, 10^{-4}, 10^{-3}, 10^{-2}\}$. Higher sparsity leads to a slight decrease in performance, lower sparsity is softer on symmetry breaking, which may reduce interpretability. This parameter has little influence as long as the sparsity remains higher than $50 \%$.

\paragraph{Word-clouds.}In \autoref{fig:nature_latent}, we form word-clouds associated with the $k$-th MSTON network $\D \l_k$ as follows. We compute the correlations between each classification map $\w_c$, associated with a condition $c$, and the network $\D \l_k$ as
\begin{equation}
    d_{k, c} = \frac{\langle \D \l_k, \w_c \rangle}{{\Vert \D \l_k \Vert}_2 {\Vert \w_c \Vert}_2}.
\end{equation}
We then show the 20 contrast names with highest correlation values---this corresponds to the contrasts whose likelihood increases the most when the input data is pushed in the direction of $\D \l_k$. The height of the contrast name $c$ in the word-cloud reflects the rank of the contrast in the sorted values $(d_{k,c})_c$ and the value $d_{k, c}$, using heuristics from the Python \textit{word\_cloud} package (\url{https://github.com/amueller/word\_cloud}).

\subsection{Validation metrics}\label{sec:metrics}

We used two metrics to measure the performance of our models.
To compare per-study decoding accuracy, we use the multi-class accuracy, defined as
\begin{equation}
    a^j = \frac{\#\{i \in [c^j n^j], \hat y^j_i = y^j_i\}}{c^j n^j},
\end{equation}
for study $j$, where $(\hat y^j_i)_{i \in [c^j n^j]}$ and  $(y^j_i)_{i \in [c^j n^j]}$
encodes the predicted and ground-truth contrasts, respectively. Box plots presented in
\autoref{fig:nature_quantitative} and Figs \ref{fig:l2}--\ref{fig:comparison_method_gm} reports
the median and 25\%, 75\% quantiles of
\begin{equation}
    \{a^j_r - \bar a^j_0, j \in [1, \dots, N], r \in [1, 2, \cdots, 20]\},
\end{equation}
where $r$ is the half-split run index and $\bar a^j_0$ is the median accuracy obtained for study $j$ over $20$ half-split.

We use balanced accuracy to measure the performance relative to a single contrast $y \in [1, \dots, c^j]$. It corresponds to the average of 1) the proportion of z-maps being correctly classified into $y$ and
2) the proportion of z-maps being correctly classified into other classes. This metric has the advantage of being comparable across studies, as its chance level is always 50\% no matter the number of contrasts in the study. We recall that the balanced accuracy $b_y^i$ for study $j$
and contrast $y$ in $[1, \dots, c^j]$ is defined as
\begin{align}
    b^j_y &\triangleq \frac{1}{2} \big( 
        \frac{n^j}{\#\{i \in [1, 2, \dots, c^j n^j], \hat y^j_i = y\}} +
        \\ &\phantom{=}
        \frac{n^j (c^j - 1)}{\#\{i \in [1, 2, \dots, c^j n^j], \hat y^j_i \neq y\}}
        \big).
\end{align}

\subsection{Quantitative results per study, task and contrast}

We report the accuracies displayed in \autoref{fig:nature_quantitative} in \autoref{table:accuracy} (multi-study decoding), and the ones displayed in \autoref{fig:per-task-quantitative} in \autoref{table:accuracy_task} (multi-task decoding). We report the list of all contrasts used in this paper in \autoref{table:all_contrasts}, as provided by the authors of each study. We report the associated balanced-accuracy when performing multi-study decoding, i.e. when we predict each contrasts among the set of all contrasts of a given study.

\renewcommand{\thetable}{\Alph{table}}

\begin{table}
    \footnotesize    
    \caption{Accuracies per study in multi-study decoding.}\label{table:accuracy}
\begin{tabular}{p{6cm}p{1cm}p{1.2cm}p{1.2cm}p{1.5cm}}
\toprule
{} & Chance level & Multi-task accuracy & Single-task accuracy & Accuracy gain \\
Study                                                                                                                                                        &              &                     &                      &               \\
\midrule
\cite{amalric2016origins} High level math \& Localizer                                                                                                       &        $3\%$ &          $83\pm1\%$ &           $81\pm1\%$ &     $2\pm1\%$ \\
\cite{pinel2007} The ARCHI project                                                                                                                           &        $3\%$ &          $87\pm1\%$ &           $86\pm1\%$ &     $1\pm1\%$ \\
\cite{papadopoulos_brainomics_2017} Brainomics                                                                                                               &        $5\%$ &          $94\pm1\%$ &           $88\pm2\%$ &     $5\pm1\%$ \\
\cite{shafto_cambridge_2014} CamCAN                                                                                                                          &       $20\%$ &          $66\pm1\%$ &           $66\pm1\%$ &     $0\pm1\%$ \\
\cite{cauvet2012traitement,hara2009neural} Music structure \& Sentence structure                                                                             &        $5\%$ &          $48\pm2\%$ &           $45\pm2\%$ &     $3\pm2\%$ \\
\cite{devauchelle2009sentence} Sentence/music complexity                                                                                                     &        $4\%$ &          $38\pm3\%$ &           $34\pm3\%$ &     $4\pm3\%$ \\
\bottomrule
\end{tabular}

\end{table}

\footnotesize
\begin{longtable}{p{1cm}p{4cm}p{1cm}p{1.2cm}p{1.2cm}p{1.5cm}}
\caption{Accuracies per task in multi-site multi task decoding.}\label{table:accuracy_task}\\
\toprule
                                 &                    & Chance level & Multi-task accuracy & Single-task accuracy & Accuracy gain \\
Study & Task &              &                     &                      &               \\
\midrule
\endfirsthead
\toprule
                                 &                    & Chance level & Multi-task accuracy & Single-task accuracy & Accuracy gain \\
Study & Task &              &                     &                      &               \\
\midrule
\endhead
\midrule
\multicolumn{6}{r}{{Continued on next page}} \\
\midrule
\endfoot

\bottomrule
\endlastfoot
\cite{amalric2016origins} & High level math &        $8\%$ &          $81\pm3\%$ &           $75\pm3\%$ &     $6\pm3\%$ \\
                                 & Localizer &        $6\%$ &          $84\pm2\%$ &           $76\pm3\%$ &     $8\pm2\%$ \\
\cite{pinel2007} & Emotion regulation &       $17\%$ &          $72\pm2\%$ &           $68\pm2\%$ &     $4\pm2\%$ \\
                                 & Localizer &        $9\%$ &          $96\pm1\%$ &           $93\pm1\%$ &     $3\pm1\%$ \\
                                 & Saccade &       $20\%$ &          $93\pm2\%$ &           $90\pm2\%$ &     $2\pm1\%$ \\
                                 & Social &       $12\%$ &          $85\pm2\%$ &           $82\pm2\%$ &     $3\pm2\%$ \\
\cite{papadopoulos_brainomics_2017} & Localizer &        $5\%$ &          $91\pm1\%$ &           $86\pm2\%$ &     $5\pm1\%$ \\
\cite{shafto_cambridge_2014} & Audio-video frequency &       $33\%$ &          $52\pm1\%$ &           $54\pm1\%$ &    $-2\pm1\%$ \\
\end{longtable}

\footnotesize
\begin{longtable}{p{1cm}p{2cm}p{3.5cm}p{1.2cm}p{1.2cm}p{1.5cm}}
\caption{List of all contrasts used in this paper, per study of origin and task.}\label{table:all_contrasts}\\
\toprule
                                 &                    &                                            & Multi-study B-acc & Voxel-level B-acc &    B-acc gain \\
Study & Task & Contrast &                   &                   &               \\
\midrule
\endfirsthead
\toprule
                                 &                    &                                            & Multi-study B-acc & Voxel-level B-acc &    B-acc gain \\
Study & Task & Contrast &                   &                   &               \\
\midrule
\endhead
\midrule
\multicolumn{6}{r}{{Continued on next page}} \\
\midrule
\endfoot

\bottomrule
\endlastfoot
\cite{amalric2016origins} & High level math & body vs baseline &        $90\pm5\%$ &        $90\pm5\%$ &     $1\pm5\%$ \\
                                 &                    & body vs checkerboard &        $94\pm5\%$ &        $97\pm3\%$ &    $-3\pm4\%$ \\
                                 &                    & checkerboard vs baseline &        $88\pm4\%$ &        $89\pm5\%$ &    $-1\pm5\%$ \\
                                 &                    & equation vs baseline &        $83\pm4\%$ &        $84\pm5\%$ &    $-1\pm4\%$ \\
                                 &                    & equation vs number &        $97\pm2\%$ &        $96\pm3\%$ &     $2\pm3\%$ \\
                                 &                    & face vs baseline &        $90\pm5\%$ &        $92\pm3\%$ &    $-2\pm4\%$ \\
                                 &                    & face vs house &       $100\pm0\%$ &        $98\pm3\%$ &     $2\pm3\%$ \\
                                 &                    & house vs baseline &        $88\pm6\%$ &        $88\pm8\%$ &     $0\pm7\%$ \\
\end{longtable}

\end{document}